\def\year{2022}\relax
\documentclass[letterpaper]{article} 
\usepackage{aaai22}  
\usepackage{times}  
\usepackage{helvet}  
\usepackage{courier}  
\usepackage[hyphens]{url}  
\usepackage{graphicx} 
\usepackage{natbib}  
\usepackage{caption} 
\DeclareCaptionStyle{ruled}{labelfont=normalfont,labelsep=colon,strut=off} 
\usepackage{algorithm}
\usepackage{algorithmic}

\usepackage{newfloat}
\usepackage{listings}
\floatstyle{ruled}
\newfloat{listing}{tb}{lst}{}
\floatname{listing}{Listing}

\title{Fourier Representations for Black-Box \\ Optimization over Categorical Variables}
\author{
    Hamid Dadkhahi\textsuperscript{\rm 1}\thanks{Work done while at IBM Research.},
    Jesus Rios\textsuperscript{\rm 2},
    Karthikeyan Shanmugam\textsuperscript{\rm 2},
    Payel Das\textsuperscript{\rm 2}
}
\affiliations{
    \textsuperscript{\rm 1} Amazon \\
    \textsuperscript{\rm 2} IBM Research AI\\
    dadkhami@amazon.com, jriosal@us.ibm.com, karthikeyan.shanmugam2@ibm.com, daspa@us.ibm.com
}

\usepackage[utf8]{inputenc} 
\usepackage{booktabs}       
\usepackage{amsfonts}       
\usepackage{nicefrac}       
\usepackage{microtype}      

\usepackage{subcaption}
\usepackage{amsmath}
\usepackage{amsthm}
\usepackage{enumitem}
\usepackage{amssymb}
\usepackage{placeins}

\newtheorem{theorem}{Theorem}[section]

\newtheorem{remark}{Remark}
\newtheorem*{exmp}{Example}

\makeatletter
\def\P(#1){\expandafter\P@i#1||\@nil}
\def\P@i#1|#2|#3\@nil{\ifx\relax#2\relax \Pr(#1)\else\Pr( #1|#2 )\fi}
\def\Pr{\mathsf{Pr}}
\makeatother

\usepackage{xspace}

\begin{document}

\maketitle

\begin{abstract}

Optimization of real-world black-box functions defined over purely categorical variables is an active area of research. In particular, optimization and design of biological sequences with specific functional or structural properties have a profound impact in medicine, materials science, and biotechnology. Standalone search algorithms, such as simulated annealing (SA) and Monte Carlo tree search (MCTS), are typically used for such optimization problems. In order to improve the performance and sample efficiency of such algorithms, we propose to use existing methods in conjunction with a surrogate model for the black-box evaluations over purely categorical variables. To this end, we present two different representations, a group-theoretic Fourier expansion and an abridged one-hot encoded Boolean Fourier expansion. To learn such representations, we consider two different settings to update our surrogate model. First, we utilize an adversarial online regression setting where Fourier characters of each representation are considered as experts and their respective coefficients are updated via an exponential weight update rule each time the black box is evaluated. Second, we consider a Bayesian setting where queries are selected via Thompson sampling and the posterior is updated via a sparse Bayesian regression model (over our proposed representation) with a regularized horseshoe prior. Numerical experiments over synthetic benchmarks as well as real-world RNA sequence optimization and design problems demonstrate the representational power of the proposed methods, which achieve competitive or superior performance compared to state-of-the-art counterparts, while improving the computation cost and/or sample efficiency, substantially.

\end{abstract}

\section{Introduction}

A plethora of practical optimization problems involve black-box functions, with no simple analytical closed forms, that can be evaluated at any arbitrary point in the domain. Optimization of such black-box functions poses a unique challenge due to restrictions on the number of possible function evaluations, as evaluating functions of real-world complex processes is often expensive and time consuming. Efficient algorithms for global optimization of expensive black-box functions take past queries into account in order to select the next query to the black-box function more intelligently. 
Black-box optimization of real-world functions defined over purely categorical type input variables has not been studied in the literature as extensively as its integer, continuous, or mixed variables counterparts.

Categorical type variables are particularly challenging when compared to integer or continuous variables, as they do not have a natural ordering. However, many real-world functions are defined over categorical variables. 
One such problem, which is of wide interest, is the design of optimal chemical or biological (protein, RNA, and DNA) molecule sequences, which are constructed using a vocabulary of fixed size, e.g. 4 for DNA/RNA. 
Designing optimal molecular sequences with novel structures or improved functionalities is of paramount importance in material science, drug and vaccine design, synthetic biology and many other applications, see \cite{dixon2010reengineering,ng2019modular, hoshika2019hachimoji, yamagami2019design}. Design of optimal sequences is a difficult black-box optimization problem over a combinatorially large search space \cite{stephens2015big}, in which function evaluations often rely on either wet-lab experiments, physics-inspired simulators, or knowledge-based computational algorithms, which are slow and expensive in practice. Another problem of interest is the constrained design problem, e.g. find a sequence given a specific structure (or property), which is inverse of the well-known folding problem discussed in \cite{dill2012protein}. This problem is complex due to the strict structural constraints  imposed on the sequence. In fact, one of the ways to represent such a complex structural constraint is to constrain the next choice sequentially based on the sequence elements that have been chosen a priori. Therefore, we divide the black box optimization problem into two settings, depending on the constraint set: $(i)$ Generic Black Box Optimization (BBO) problem referring to the unconstrained case and $(ii)$ Design Problem that refers to the case with complex sequential constraints.

Let $x_t$ be the $t$-th sequence evaluated by the black box function $f$. The key question in both settings is the following: Given prior queries $x_1,x_2 \ldots x_t$ and their evaluations $f(x_1) \ldots f(x_t)$, \textit{how to choose the next query $x_{t+1}$}? This selection must be devised so that over a finite budget of black-box evaluations, one is closest to the minimizer in an expected sense over the acquisition randomness.

In the literature, for design problems with sequential constraints, MCTS (Monte Carlo Tree Search) based search algorithms are often used with black box function evaluations. In the generic BBO problems in the unconstrained scenario, Simulated Annealing (SA) based techniques are typically used as search algorithms. A key \textit{missing} ingredient in the categorical domain is a surrogate model for the black-box evaluations that can interpolate between such evaluations and facilitate \textit{cost-free} approximate evaluations from the surrogate model internally in order to reduce the need for frequently accessing real black-box evaluations,  
thus leading to improved sample efficiency in both generic and design black-box optimization problems. 
Due to the lack of efficient interpolators in the categorical domains, existing search algorithms suffer under constrained computational budgets, due to reliance on only real black-box evaluations.




\textbf{Contributions}: We address the above problem in our work. Our main contributions are as follows:
\vspace{-1mm}
\begin{enumerate}[leftmargin=5mm, itemsep=-1pt]
    \item We present two representations for modeling real-valued combinatorial functions over categorical variables, which we then use in order to learn a surrogate model for the generic BBO problem and the design problem, with a finite budget on the number of queries. The abridged one-hot encoded Boolean Fourier representation is novel to this work. The use of group-theoretic Fourier representation for modeling functions over categorical variables, and in particular their optimization, is novel to this work.
    \item Numerical experiments, over synthetic benchmarks as well as real-world biological (RNA) sequence optimization demonstrate the competitive or superior performance of the proposed methods incorporating our representations over state-of-the-art counterparts, while substantially reducing the computation time. We further evaluate the performance of our algorithms in design problems (inverse folding problem) and demonstrate the superior performance of our methods in terms of sample efficiency over the state-of-the-art counterparts.
\end{enumerate}

\section{Related Work}

\textbf{Black-Box Optimization}: Hutter et al. \cite{SMAC2011} suggest a surrogate model based on random forests to address optimization problems over categorical variables. The proposed SMAC algorithm uses a randomized local search under the expected improvement acquisition criterion to obtain candidate points for black-box evaluations. Bergstra et al. \cite{TPE2011} suggest a tree-structured Parzen estimator (TPE) for approximating the surrogate model, and maximizes the expected improvement criterion to find candidate points for evaluation.
For optimization problems over Boolean variables, multilinear polynomials \cite{BOCS, COMEX} and Walsh functions \cite{Walsh2019, MercerFeatures} have been used in the literature. 

Bayesian Optimization (BO) is a commonly used approach for optimization of black-box functions \cite{shahriari2015taking}. 
However, limited work has addressed incorporation of categorical variables in BO. 
Early attempts were based on converting the black-box optimization problem over categorical variables to that of  continuous variables 
\cite{gomez2018automatic, golovin2017google, garrido2020dealing}.
A few BO algorithms have been specifically designed for black-box functions over combinatorial domains. In particular, the BOCS algorithm \cite{BOCS}, primarily devised for Boolean functions, employs a sparse monomial representation to model the interactions among different variables, and uses a sparse Bayesian linear regression method to learn the model coefficients. 
The COMBO algorithm of \cite{COMBO} uses Graph Fourier Transform over a combinatorial graph, constructed via graph Cartesian product of variable subgraphs, to gauge the smoothness of the black-box function. 
However, both BOCS and COMBO are hindered by associated high computation costs, which grow polynomially with both the number of variables and the number of function evaluations. 
More recently, a computationally efficient black-box optimization algorithm (COMEX) \cite{COMEX} was introduced to address the computational impediments of its Bayesian counterparts. COMEX adopts a Boolean Fourier representation as its surrogate model, which is updated via an exponential weight update rule. 
Nevertheless, COMEX is limited to functions over the Boolean hypercube. 

We generalize COMEX to handle functions over categorical variables by proposing two representations for modeling functions over categorical variables: an abridged one-hot encoded Boolean Fourier representation and Fourier representation on finite Abelian groups. The utilization of the latter representation as a surrogate model in combinatorial optimization algorithms is novel to this work. Factorizations based on (vanilla) one-hot encoding has been previously 
(albeit briefly) 
suggested in \cite{BOCS} to enable black-box optimization algorithms designed for Boolean variables to address problems over categorical variables. Different from \cite{BOCS}, we show that we can significantly reduce the number of additional variables introduced upon one-hot encoding, and that such a reduced representation is in fact complete and unique. We incorporate the latter representation to extend a modified version of BOCS to problems over categorical variables, efficiently.


\textbf{Sequence Design}: For design problems, we focus on the RNA sequence design problem (RNA inverse folding). The goal is to find an RNA sequence consistent with a given secondary structure, as the functional state of the RNA molecule is determined by the latter structure \cite{hofacker1994fast}. Earlier RNA design methods explore the search space by trial and error and use classic cost function minimization approaches such as adaptive random walk \cite{hofacker2003vienna}, probabilistic sampling \cite{zadeh2011nupack}, and genetic algorithms \cite{taneda2015multi}. Recent efforts employ more advanced machine learning methods such as 
different Monte Carlo Tree Search (MCTS) algorithms,  e.g. MCTS-RNA \cite{MCTSRNA} or Nested MCTS \cite{portela2018unexpectedly}, and reinforcement learning that either performs a local search as in \cite{eastman2018solving} or learns complete candidate solutions from scratch \cite{runge2018learning}. In all these approaches, the assumption is that the algorithm has access to a large number of function evaluations, whereas we are interested in sample efficiency of each algorithm. 

\textbf{Learning to Search}: As an alternative to parameter-free search methods (such as SA), \cite{Amortized2020} suggests to use a parameterized policy to generate candidates that maximize the acquisition function in Bayesian optimization over discrete search spaces. Our MCTS-based algorithm is similar in concept to \cite{Amortized2020} in the sense that the tabular value functions are constructed and maintained over different time steps. However, we are maintaining value functions rather than a policy network. Finally, while \cite{Amortized2020} trains a policy network, \cite{rank_learning} uses a rank learning algorithm to search.

\textbf{Meta-heuristics}: A variety of discrete search algorithms and meta-heuristics have been studied in the literature for combinatorial optimization over categorical variables. Such algorithms, including Genetic Algorithms \cite{holland1978cognitive}, Simulated Annealing \cite{SA}, and Particle Swarms \cite{488968}, are generally inefficient in finding the global minima. In the context of biological sequence optimization, the most popular method is directed evolution \cite{arnold1998design}, which explores the space by only making small mutations to existing sequences. In the context of sequence optimization, a recent promising approach consists of fitting a neural network model to predict the black box function and then applying gradient ascent on the latter model \cite{killoran2017generating, bogard2019deep, liu2020antibody}. This approach allows for a continuous relaxation of the discrete search space making possible step-wise local improvements to the whole sequence at once based on a gradient direction. However, these methods have been shown to suffer from vanishing gradients \cite{Linder2020}. Further, the projected sequences in the continuous relaxation space may not be recognized by the predictors, leading to poor convergence. 
Generative model-based optimization approaches aim to learn distributions whose expectation coincides with evaluations of the black box and try to maximize such expectation \cite{gupta2019feedback, brookes2019conditioning}. However, such approaches require a pre-trained generative model for optimization. 

\textbf{Latent Space Optimization}: The key idea is to employ an encoder-decoder architecture to learn a continuous representation from the data and perform BO in a latent space \cite{gomez2018automatic, Tripp2020sample}. This approach has two main drawbacks: it could generate a large fraction of invalid structures, and  it requires a large database of relevant structures, for training an auto-encoder, which may not be available in many applications with scarce data availability.

\textbf{Variable-Length Sequence Optimization}: Moss et al. \cite{BOSS2020} propose an algorithm called BOSS for optimization over strings of variable lengths. They use a Gaussian process surrogate model based on string kernels and perform acquisition function maximization for spaces with syntactic constraints. On the contrary, our main focus is on fixed length generic BBO and design problems.

\textbf{Ensemble (Population-based) Methods}: Angermueller et al. \cite{population_based} propose a population-based approach which generates batches of sequences by sampling from an ensemble of methods, rather than from a single method. This sampling is carried out proportional to the quality of sequences that each method has found in the past. A similar ensemble concept has been used in \cite{sinai2021adalead}. In our work, we focus on individual methods; however, we add that since we are proposing two distinct representations employed via different algorithms (online and Bayesian), and the performance of each varies across different tasks, it would be interesting to consider an ensemble method over different algorithms as a future direction.

\textbf{Batch (Offline) Optimization}: Batch optimization techniques \cite{Angermueller2020Model, brookes2019conditioning, Fannjiang2020} are a related but fundamentally distinct line of work, where black-box evaluations are carried out in sample batches rather than in an active/sequential manner. To highlight the distinction between the two settings we iterate our problem setting: Given query access to a combinatorial black box (i.e. it can be queried on a single point), what is the best solution we can obtain in a given number $T$ of such queries. One wishes to minimize the number of queries and hence goes for a highly adaptive sequential algorithm.

\section{Black-Box Optimization over Categorical Variables}
\label{sec:categorical}

\textbf{Problem Setting}:
Given the combinatorial categorical domain $\mathcal{X} = [k]^n$ and a constraint set ${\cal C} \subseteq \mathcal{X}$, with $n$ variables each of cardinality $k$, the objective is to find 
\begin{equation}
	x^* = \arg\min_{x \in \mathcal{C}} f(x)
\end{equation}
where $f: \mathcal{X} \mapsto \mathbb{R}$ is a real-valued combinatorial function. We assume that $f$ is a black-box function, which is computationally expensive to evaluate. As such, we are interested in finding $x^*$ in as few evaluations as possible. We consider two variations of the problem depending on how the constraint set ${\cal C}$ is specified.

\textbf{Generic BBO Problem:} In this case, the constraint set ${\cal C}={\cal X}$. For example, RNA sequence optimization problem that searches for an RNA sequence with a specific  property optimized lies within this category. A score for every RNA sequence, reflecting the property we wish to optimize, is evaluated by a black box function.

\textbf{Design Problem:} The constraint set is complex and is only sequentially specified. For every sequence of $x_1x_2 \ldots x_i$ consisting of $i$ characters from the alphabet $[k]$, the choice of the next character $x_{i+1} \in {\cal C}(x_1x_2 \ldots x_{i}) \subseteq [k]$ is specified by a constraint set function ${\cal C}(x_1 \ldots x_i)$. The RNA inverse folding problem in \cite{runge2018learning} falls into this category, where the constraints on the RNA sequence are determined by the sequential choices one makes during the sequence design. The goal is to find the sequence that is optimal with respect to a pre-specified structure that also obeys complex sequential constraints.

\textbf{Proposed Techniques}:
To address this problem, we adopt a surrogate model-based learning framework as follows. The surrogate model (to represent the black-box function) is updated sequentially via black-box function evaluations observed until time step $t$. An acquisition function is then obtained from the surrogate model. The selection of candidate points for black-box function evaluation is carried out via an acquisition function optimizer (AFO), which uses the acquisition function as an inexpensive proxy (to make many internal calls) for the black-box function and produces the next candidate point to be evaluated.
We consider this problem in two different settings: online and Bayesian. The difference between the two settings is that in the former the acquisition function is exactly our surrogate, whereas in the latter the acquisition function is obtained by sampling from the posterior distribution of the surrogate.



In the sequel, we propose two representations that can be used in surrogate models for black-box combinatorial functions over categorical variables. These representations serve as direct generalizations of the Boolean surrogate model based on Fourier expansion proposed in \cite{COMEX, BOCS} in the sense that they reduce to the Fourier representation for real-valued Boolean functions when the cardinality of the categorical variables is two. In addition, both  approaches can be modified to address the more general case where different variables are of different cardinalities. However, for ease of exposition, we assume that all the variables are of the same cardinality. Finally, we introduce two popular search algorithms to be used as acquisition function optimizers in conjunction with the proposed surrogate models in order to select new queries for subsequent black-box function evaluations.

\subsection{Representations for  the Surrogate Model}
\label{sec:model}
We present two representations for combinatorial functions $f:[k]^n \rightarrow \mathbb{R}$ and an algorithm to update from the black-box evaluations. The representations use the Fourier basis in two different ways.

\textbf{Abridged One-Hot Encoded Boolean Fourier Representation}: 
The one-hot encoding of each variable $x_i: i \in [n]$ can be expressed as a $(k-1)$-tuple $(x_{i1}, x_{i2}, \ldots, x_{i(k-1)})$, where $x_{ij} \in \{-1, 1\}$ are Boolean variables with the constraint that at most one such variable can be equal to $-1$ for any given $x_i \in [k]$.

We consider the following representation for the combinatorial function $f$:
\begin{equation}
	f_{\alpha}(x) = \sum_{m=0}^n \sum_{\mathcal{I} \in \binom{[n]}{m}} \sum_{\mathcal{J} \in [k-1]^{|\mathcal{I}|}} \alpha_{\mathcal{I}, \mathcal{J}} \psi_{\mathcal{I}, \mathcal{J}}(x)
\label{eq:factorization}
\end{equation}
where $[k-1]^{|\mathcal{I}|}$ denotes $|\mathcal{I}|$-fold cartesian product of the set $[k-1] = \{1, 2, \ldots, k - 1\}$, $\binom{[n]}{m}$ designates the set of $m$-subsets of the set $[n]$, and the monomials $\psi_{\mathcal{I}, \mathcal{J}}(x)$ can be written as
\begin{equation}
	\psi_{\mathcal{I}, \mathcal{J}}(x) = \prod_{\{(i, j): i = \mathcal{I}_{\ell}, j = \mathcal{J}_{\ell}, \ell \in [|\mathcal{I}|]\}} x_{ij}
\label{eq:monomials}
\end{equation}

A second order approximation (i.e. at $m = 2$) of the representation in \eqref{eq:factorization} can be expanded in the following way:
\begin{align}
	\widehat{f}_{\alpha}(x) &= \alpha_0 + \sum_{i \in [n]} \sum_{\ell \in [k-1]} \alpha_{i \ell} x_{i \ell} \nonumber \\
	&+ \sum_{(i, j) \in \binom{[n]}{2}} \sum_{(p, q) \in [k-1]^2} \alpha_{ijpq} x_{ip} x_{jq}.
\label{eq:simple_factorization}
\end{align}

\begin{exmp}
For $n = 2$ variables $x_1$ and $x_2$, each of which with cardinality $k = 3$, we have the one-hot encoding of $(x_{11}, x_{12})$ and $(x_{21}, x_{22})$ respectively. From Equation \eqref{eq:simple_factorization}, the one-hot encoding factorization for this example can be written as
\begin{align*}
	f(x) &= \alpha_0 + \alpha_1 x_{11} + \alpha_2 x_{12} + \alpha_3 x_{21} + \alpha_4 x_{22} \\ 
	&+ \alpha_5 x_{11} x_{21} + \alpha_6 x_{11} x_{22} + \alpha_7 x_{12} x_{21} + \alpha_8 x_{12} x_{22}.
\end{align*}
\end{exmp}

Note that the representation in Equation \eqref{eq:factorization} has far less terms than a vanilla one-hot encoding with all the combinations of one-hot variables included (as suggested in \cite{BOCS}). The reason for this reduction is two-fold: $(i)$ $(k-1)$ Boolean variables model each categorical variable of cardinality $k$, and more importantly $(ii)$ each monomial term has at most one Boolean variable $x_{ij}$ from its corresponding parent categorical variable $x_i$ (See the Appendix for the exact quantification of this reduction). The following theorem states that this reduced representation is in fact unique and complete.

\begin{theorem}
\label{thm:factorization}
The representation in Equation \eqref{eq:factorization} is complete and unique for any real-valued combinatorial function.
\end{theorem}
\begin{proof}
See Appendix \ref{app:proofs}.
\end{proof}





\textbf{Fourier Representation on Finite Abelian Groups}:
We define a cyclic group structure $\mathbb{Z}/k_i\mathbb{Z}$ over the elements of each categorical variable $x_i$ ($i \in [n]$), where $k_i$ is the cardinality of the latter variable. From the fundamental theorem of abelian groups \cite{terras1999}, there exists an abelian group $G$ which is isomorphic to the direct sum (a.k.a direct product) of the cyclic groups $\mathbb{Z}/k_i\mathbb{Z}$ corresponding to the $n$ categorical variables: 
\begin{equation}
	G \cong  \mathbb{Z}/k_1\mathbb{Z} \oplus \mathbb{Z}/k_2\mathbb{Z} \oplus \ldots \oplus \mathbb{Z}/k_n\mathbb{Z}
\end{equation}
where the latter group consists of all vectors $(a_1, a_2, \ldots, a_n)$ such that $a_i \in \mathbb{Z}/k_i\mathbb{Z}$ and $\cong$ denotes group isomorphism. We assume that $k_i=k$ ($\forall i \in [n]$) for simplicity, but the following representation could be easily generalized to the case of arbitrary cardinalities for different variables.

The Fourier representation of any complex-valued function $f(x)$ on the finite abelian group $G$ is given by \cite{terras1999}
\begin{equation}
	f(x) = \sum_{\mathcal{I} \in [k]^n} \alpha_{\mathcal{I}} \psi_{\mathcal{I}}(x)
\label{eq:group_fourier_complex}
\end{equation}
where $\alpha_{\mathcal{I}}$ are (in general complex) Fourier coefficients, $[k]^n$ is the $n$-fold cartesian product of the set $[k]$ and $\psi_{\mathcal{I}}(x)$ are complex exponentials \footnote{Note that in the general case of different cardinalities for different variables, $\mathcal{I} \in [k_1] \times [k_2] \times \ldots \times [k_n]$ where $\times$ denotes the cartesian product and the exponent denominator in the complex exponential character is replaced by $k = \mathsf{LCM}(k_1, k_2, \ldots, k_n)$.} ($k$-th roots of unity) given by
\begin{equation*}
	\psi_{\mathcal{I}}(x) = \exp( \nicefrac{2 \pi j \langle x, \mathcal{I} \rangle}{k}).
\end{equation*}
Note that the latter complex exponentials are the \textit{characters} of the representation, and reduce to the \textit{monomials} (i.e. in $\{-1, 1\}$) when the cardinality of each variable is two. A second order approximation of the representation in \eqref{eq:group_fourier_complex} can be written as:
\begin{align}
	\widehat{f}_{\alpha}(x) &= \alpha_0 + \sum_{i \in [n]} \sum_{\ell \in [k-1]} \alpha_{i \ell} \exp( \nicefrac{2 \pi j x_i \ell}{k}) \\ 
	&+ \sum_{(i, j) \in \binom{[n]}{2}} \sum_{(p, q) \in [k-1]^2} \alpha_{ijpq} \exp( \nicefrac{2 \pi j (x_i p + x_j q)}{k}). \nonumber
\end{align}
For a real-valued function $f_{\alpha}(x)$ (which is of interest here), the representation in \eqref{eq:group_fourier_complex} reduces to
\begin{align}
	f_{\alpha}(x) &= \Re\bigg\{\sum_{\mathcal{I} \in [k]^n} \alpha_{\mathcal{I}} \psi_{\mathcal{I}}(x)\bigg\} \nonumber \\ 
	&= \sum_{\mathcal{I} \in [k]^n} \alpha_{r, \mathcal{I}} \psi_{r, \mathcal{I}}(x) - \sum_{\mathcal{I} \in [k]^n} \alpha_{i, \mathcal{I}} \psi_{i, \mathcal{I}}(x)
\label{eq:group_fourier}
\end{align}
where
\begin{align}
	\psi_{r, \mathcal{I}}(x) = \cos( \nicefrac{2 \pi \langle x, \mathcal{I} \rangle}{k}) \quad & \texttt{and} \quad
	\psi_{i, \mathcal{I}}(x) = \sin( \nicefrac{2 \pi \langle x, \mathcal{I} \rangle}{k}) \nonumber \\
	\alpha_{r, \mathcal{I}} = \Re \{ \alpha_{\mathcal{I}} \} \quad 
	& \texttt{and} \quad 
	\alpha_{i, \mathcal{I}} = \Im \{ \alpha_{\mathcal{I}} \}
\label{eq:characters}
\end{align}
See the Appendix for the proof of this representation. We note that the number of characters utilized in this representation is almost twice as many as that of monomials used in the previous representation.


\subsection{Surrogate Model Learning}

\textbf{Sparse Online Regression}: We adopt the learning algorithm of combinatorial optimization with expert advice \cite{COMEX} in the following way.
We consider the monomials $\psi_{\mathcal{I}, \mathcal{J}}(x)$ in \eqref{eq:monomials} and the characters $\psi_{\ell, \mathcal{I}}(x)$ in \eqref{eq:characters} as experts. For each surrogate model, we maintain a pool of such experts, the coefficients of which are refreshed sequentially via an exponential weight update rule. 
We refer to the proposed algorithm as \textit{Expert-Based Categorical Optimization} (ECO) and the two versions of the algorithm with the two proposed surrogate models are called ECO-F (based on the One-Hot Encoded Boolean Fourier Representation) and ECO-G (based on Fourier Representation on Finite Abelian Groups), respectively. See Appendix for more details.


\textbf{Sparse Bayesian Regression}: In order to incorporate the uncertainty of the surrogate model in acquisition function for exploration-exploitation trade-off, we adopt a Bayesian setting based on Thompson sampling (TS) \cite{Thompson1933, Thompson1935}. In particular, we model the surrogate via a sparse Bayesian regression model with a regularized horseshoe prior \cite{finnish_horseshoe}. At each time step, we learn the posterior using the data observed so far, and draw a sample for the coefficients $\alpha$ from the posterior. We use the No-U-Turn Sampler (NUTS) \cite{NUTS} in order to sample the coefficients efficiently. See Appendix B for more details. Our algorithm pursues the framework of BOCS \cite{BOCS}, with the following differences for improved performance and computational efficiency: BOCS ($i$) employs a horseshoe prior \cite{horseshoe, horseshoe_sampler} and ($ii$) uses a Gibbs sampler. Experimentally, the regularized horseshoe prior outperforms the horseshoe prior, and the NUTS sampler leads to a significant speed-up in computations (e.g. a $10\times$ speed-up for the RNA optimization problem), which is critical due to the higher dimensionality of the categorical domains. We use the proposed abridged one-hot encoded Fourier representation in conjunction with this model, due to its lower number of coefficients leading to a more manageable computation cost. We refer to this algorithm as \textit{TS-based Categorical Optimization} (TCO-F).

\subsection{Acquisition Function Optimizers}
In this subsection, we discuss how two popular search algorithms, namely Simulated Annealing (SA) and Monte Carlo Tree Search (MCTS), can be utilized as acquisition function optimizers (AFO) in conjunction with a surrogate model and use cost-free evaluations of the surrogate model to select the next query for the black box evaluation. In the literature, SA has been used for the generic BBO problems, whereas MCTS has been used for the design problems.

\textbf{SA as AFO:} Our AFO is devised so as to minimize $\widehat{f}_{\alpha}(x)$, the current estimate for the surrogate model. A simple strategy to minimize $\widehat{f}_{\alpha}(x)$ is to iteratively switch each variable into the value that minimizes $\widehat{f}_{\alpha}$ given the values of all the remaining variables, until no more changes occur after a sweep through all the variables $x_i$ ($\forall i \in [n]$). A strategy to escape local minima in this context \cite{pincus1970letter} is to allow for occasional increases in $\widehat{f}_{\alpha}$ by generating a Markov Chain (MC) sample sequence (for $x$), whose stationary distribution is proportional to $ \exp(-\nicefrac{\widehat{f}_{\alpha}(x)}{s})$, where $s$ is gradually reduced to $0$. This optimization strategy was first applied by \cite{Kirkpatrick1983} in their Simulated Annealing algorithm to solve combinatorial optimization problems. We use the Gibbs sampler \cite{geman1984stochastic} to generate such an MC by sampling from the full conditional distribution of the stationary distribution, which in our case is given by the softmax distribution over $\{-\widehat{f}_{\alpha}(x_i=\ell, x_{-i})/s \}_{\ell \in [k]}$, for each  variable $x_i$ conditional on the values of the remaining variables $x_{-i}$. By decreasing $s$ from a high value to a low one, we allow the MC to first search at a coarse level avoiding being trapped in local minima. 

Algorithm \ref{algo:SA} presents our simulated annealing (SA) version for categorical domains, where $s(t)$ is an annealing schedule, which is a non-increasing function of $t$. We use the annealing schedule suggested in \cite{SA}, which follows an exponential decay with parameter $\ell$ given by $s(t) = \exp(-\nicefrac{\ell t}{n})$. In each step of the algorithm, we pick a variable $x_i$ ($i \in [n]$) uniformly at random, evaluate the surrogate model (possibly in parallel) $k$ times, once for each categorical value $\ell \in [k]$ for the chosen variable $x_i$ given the current values $x_{-i}$ for the remaining variables. We then update $x_i$ with the sampled value in $[k]$ from the corresponding softmax distribution.  


\begin{algorithm}
\caption{SA for Categorical Variables with Surrogate Model}
\begin{algorithmic}[1]
\STATE \textbf{Inputs:} surrogate model $\widehat{f}_{\alpha}$, annealing schedule $s(t)$, categorical domain $\mathcal{X}$
\STATE Initialize $x \in \mathcal{X}$
\STATE $t = 0$
\REPEAT
\STATE $i \sim \texttt{unif}([n])$
\STATE $x_i|x_{-i}\sim \texttt{Softmax}\big(\{-\nicefrac{\widehat{f}_{\alpha_t}(x_i=\ell, x_{-i})}{s(t)}\}_{\ell \in [k]}\big)$
\STATE $t \gets t + 1$
\UNTIL{Stopping Criteria}
\RETURN $x$
\end{algorithmic}
\label{algo:SA}
\end{algorithm}

\begin{algorithm}
\caption{MCTS with Surrogate Reward}
\begin{algorithmic}[1]
\STATE \textbf{Inputs:} surrogate model $\widehat{f}_{\alpha}$, search tree $\mathcal{T} $
\STATE Initialize $s_{\texttt{best}}=\{\}$,  $r_{\texttt{best}}=-\infty$ 
\REPEAT  
\STATE $s_{\texttt{leaf}} \gets \texttt{Selection}(\pi^{\mathcal{T}})$
\STATE $\mathcal{T} \gets \mathcal{T} \cup \{s_{\texttt{leaf}}\}$
\STATE $s_t \gets \texttt{Simulation}(\pi^{RS}, s_{\texttt{leaf}})$
\STATE $r \gets - \widehat{f}_{\alpha}(s_t)$
\STATE $\texttt{Backup}(s_{\texttt{leaf}}, r)$
\IF{$r > r_{\texttt{best}}$} 
\STATE $r_{\texttt{best}} \gets r \: \texttt{and} \:  s_{\texttt{best}} \gets s_t$
\ENDIF
\UNTIL{Stopping Criteria}
\RETURN $s_{\texttt{best}}$
\end{algorithmic}
\label{algo:MCTS}
\end{algorithm}

\textbf{MCTS as AFO:} We formulate the design problem as an undiscounted Markov decision process $(\mathcal{S}, \mathcal{A}, T, R)$. Each state $s \in \mathcal{S}$ corresponds to a partial or full sequence of categorical variables of lengths in $[0, n]$. The process in each episode starts with an empty sequence $s_0$, the initial state. Actions are selected from the set of permissible additions to the current state (sequence) $s_t$ at each time step $t$, $\mathcal{A}_t = \mathcal{A}(s_t) \subset \mathcal{A}$. The transition function $T$ is deterministic, and defines the sequence obtained from the juxtaposition of the current state $s_t$ with the action $a_t$, i.e. $s_{t+1} = T(s_t, a_t) = s_t \circ a_t$. The transitions leading to incomplete sequences yield a reward of zero. Complete sequences are considered as terminal states, from which no further transitions (juxtapositions) can be made. Once the sequence is complete (i.e. at a terminal state), the reward is obtained from the current surrogate reward model $\widehat{f}_{\alpha}$. Thus, the reward function is defined as $ R(s_t,a_t,s_{t+1})= - \widehat{f}_{\alpha}(s_{t+1})$ if $s_{t+1}$ is terminal, and zero otherwise. 

MCTS is a popular search algorithm used for design problems. MCTS is a rollout algorithm which keeps track of the value estimates obtained via Monte Carlo simulations in order to progressively make better selections. The UCT selection criteria, see \cite{UCT}, is typically used as tree policy, where action $a_t$ at state $s_t$ in the search tree is selected via: $\pi^{\mathcal{T}}(s_t) = \arg\max_{a \in \mathcal{A}(s_t)} Q(s_t, a) + c \sqrt{\nicefrac{\ln{N(s_t)}}{N(s_t, a)}}$, where $\mathcal{T}$ is the search tree, $c$ is the exploration parameter, $Q(s, a)$ is the state-action value estimate, and $N(s)$ and $N(s, a)$ are the visit counts for the parent state node and the candidate state-action edge, respectively. For the selection of actions in states outside the search tree, a random default policy is used: $\pi^{RS}(s_t) = \texttt{unif}(\mathcal{A}_t)$.

A summary of the proposed algorithm is given in Algorithm \ref{algo:MCTS}. Starting with an empty sequence $s_0$ at the root of the tree, we follow the tree policy until a leaf node of the search tree is reached (selection step). At this point, we append the state corresponding to the leaf node to the tree and initialize a value function estimate for its children (extension step). From the reached leaf node we follow the default policy until a terminal state is reached. At this point, we plug the sequence corresponding to this terminal state into the surrogate reward model $-\widehat{f}_{\alpha}$ and observe the reward $r$. This reward is backed up from the leaf node to the root of the tree in order to update the value estimates $Q(s, a)$ via Monte Carlo (i.e. using the average reward) for all visited $(s, a)$ pairs along the path. This process is repeated until a stopping criterion (typically a max number of playouts) is met, at which point the sequence $s_{\texttt{best}}$ with maximum reward $r_{\texttt{best}}$ is returned as the output of the algorithm. 

\textbf{Computational Complexity}: 
The computational complexity per time step associated with learning the surrogate model via the Hedge algorithm, for both representations introduced in \ref{sec:model}, is in $\mathcal{O}(d) = \mathcal{O}(k^{m-1} n^m)$, and is thus linear in the number of experts $d$. Moreover, the complexity of Algorithm \ref{algo:SA} is in $\mathcal{O}(k k^{m-1} n^{m-1} n) = \mathcal{O}(k d)$, assuming that the number of iterations in each SA run is in $\mathcal{O}(n)$. Hence, the overall complexity of the algorithm is in $\mathcal{O}(k d)$.
Finally, the computational complexity of each playout in Algorithm \ref{algo:MCTS} is in $\mathcal{O}(kn)$, leading to an overall complexity of $\mathcal{O}(k d)$, assuming $\mathcal{O}(\tfrac{d}{n})$ playouts per time step.
On the other hand, when Bayesian regression is used to learn the surrogate model, NUTS dominates the computation time with a complexity of $\mathcal{O}(d^{\nicefrac{5}{4}})$ per independent sample.

\section{Experiments and Results}
\label{sec:experiments}

In this section, we measure the performance of the proposed representations, when used as surrogate/reward model in conjunction with search algorithms (SA and MCTS) in BBO and design problems. The learning rate used in exponential weight updates is selected via the anytime learning rate schedule suggested in \cite{COMEX} and \cite{adaptive_EG} (see Appendix \ref{app:descriptions}). The maximum degree of interactions used in our surrogate models is set to two for all the problems; increasing the max order improved the results only marginally (see Appendix \ref{app:order}). The sparsity parameter $\lambda$ in exponential weight updates is set to $1$ in all the experiments following the same choice made in \cite{COMEX}. Experimentally, the learning algorithm is fairly insensitive to the variations in the latter parameter. In each experiment, we report the results averaged over multiple runs ($20$ runs in BBO experiments; $10$ runs in design experiments) $\pm$ one standard error of the mean. 
The experiments are run on machines with CPU cores from the Intel Xeon E5-2600 v3 family.

\textbf{BBO Experiments}:
We compare the performance of our ECO/TCO algorithms in conjunction with SA with two baselines, random search (RS) and simulated annealing (SA), as well as a state-of-the-art Bayesian combinatorial optimization algorithm (COMBO) \cite{COMBO}. In particular, we consider two synthetic benchmarks (Latin square problem and pest control problem) and a real-word problem in biology (RNA sequence optimization). In addition to the performance of the algorithms in terms of the best value of $f(x)$ observed until a given time step $t$, we measure the average computation time per time step of our algorithm versus that of COMBO. 
The decay parameter used in the annealing schedule of SA is set to $\ell = 3$ in all the experiments. In addition, the number of SA iterations $T$ is set to $3 \times n$ and $6 \times n$ for ECO and TCO, respectively. Intuitively, for the ECO algorithm, each of these parameters creates an exploration-exploitation trade-off. The smaller (larger) the value of $\ell$ or $T$, the more exploratory (exploitative) is the behavior of SA. The selected values seem to create a reasonable balance; tuning these parameters may improve the performance of the algorithm. On the other hand, the TCO algorithm has an exploration mechanism built into it via Thompson sampling; hence, we use a higher number of SA iterations to maximize exploitation in AFO.

\textbf{Synthetic Benchmarks}: We consider two synthetic problems: Latin square problem \cite{handbook_combinatorial}, a commonly used combinatorial optimization benchmark, and the pest control problem considered in \cite{COMBO} (see Appendix for the latter results). In both problems, we have $n = 25$ categorical variables, each of cardinality $k = 5$.
A Latin square of order $k$ is a $k \times k$ matrix of elements $x_{ij} \in [k]$, such that each number appears in each row and column exactly once. When $k = 5$, the problem of finding a Latin square has $161,280$ solutions in a space of dimensionality $5^{25}$. We formulate the problem of finding a Latin square of order $k$ as a black-box optimization by imposing an additive penalty of $1$ for any repetition of numbers in any row or column. Hence, function evaluations are in the range $[0, 2 k (k-1)]$, and a function evaluation of $0$ corresponds to a Latin square of order $k$. We consider a noisy version of this problem, where an additive Gaussian noise with $0$ mean and standard deviation of $0.1$ is added to function evaluations observed by each algorithm. 

\begin{figure}[ht]
\center
\includegraphics[width=.9\linewidth]{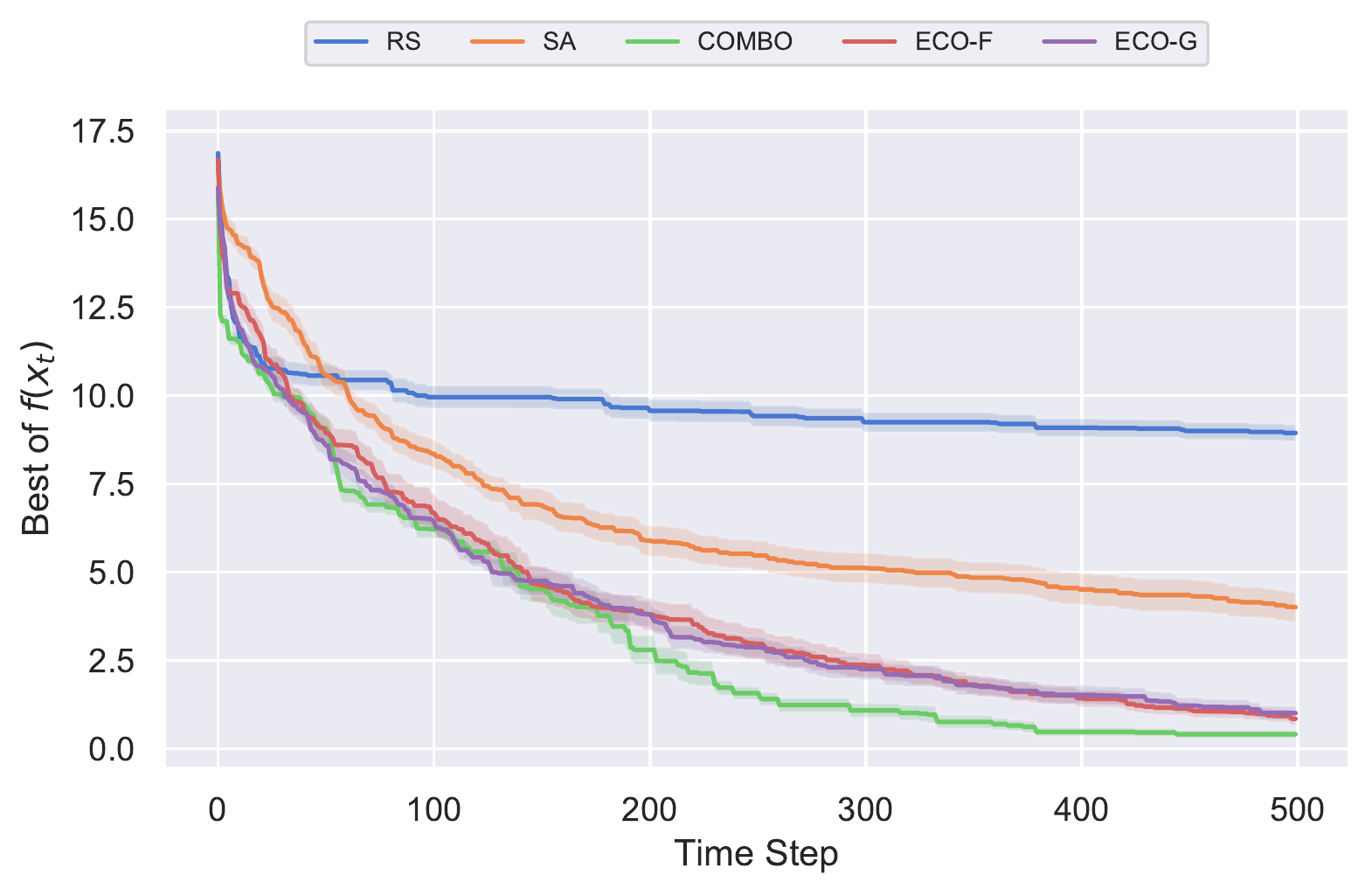} 
\caption{Latin Square problem with $n = 25$.
}
\label{fig:latin_square_main}
\vspace{-1.5mm}
\end{figure}

Figure \ref{fig:latin_square_main} demonstrates the performance of different algorithms, in terms of the best function value found until time $t$, over $500$ time steps. Both ECO-F and ECO-G outperform the baselines with a considerable margin. In addition, ECO-G outperforms COMBO at $130$ samples. At larger time steps, COMBO outperforms the other algorithms; however, this performance comes at the price of a far larger computation time. As demonstrated in Table \ref{table:times_main}, ECO-F and ECO-G offer a speed-up of roughly $100$ and $50$, resp., over COMBO. We note that TCO-F (not included in the plot) performs poorly (similar to RS) on this problem, which can be attributed to the strong promotion of sparsity by the regularized horseshoe prior and the fact that the Latin Square problem has a dense representation (we observed a similar behavior from the horseshoe prior of \cite{BOCS}).

\begin{table}[t]
\caption{Average computation time per step (in Seconds).}
\label{table:times_main}
\begin{center}
\begin{small}
\begin{sc}
\begin{tabular}{lccccccr}
\toprule
Data & Latin Square & Pest Cont. & RNA Opt. \\
\midrule
COMBO & 170.4 & 151.0 & 253.8 \\
ECO-F & 1.5 & 1.4 & 2.0 \\
ECO-G & 3.6 & 3.3 & 5.7  \\
TCO-F & 55.7 & 53.2 & 67.0  \\
\bottomrule
\end{tabular}
\end{sc}
\end{small}
\end{center}
\end{table}

\textbf{RNA Sequence Optimization Problem}: 
Consider an RNA sequence as a string $A=a_1 \ldots a_n$ of $n$ letters (nucleotides) over the alphabet $\Sigma=\{A, U, G, C\}$. A pair of complementary nucleotides $a_i$ and $a_j$, where $i<j$, can interact with each other and form a base pair (denoted by $(i, j)$), A-U, C-G and G-U being the energetically stable pairs. Thus, the secondary structure, i.e. the minimum free energy structure, of an RNA can be represented by an ensemble of pairing bases. 
A number of RNA folding algorithms \cite{ViennaRNA, markham2008unafold} use a thermodynamic model (e.g. \cite{zuker1981optimal}) and dynamic programming to estimate MFE of a sequence. However, the $\mathcal{O}(n^3)$ time complexity of these algorithms prohibits their use for evaluating substantial numbers of RNA sequences \cite{gould2014computational} and exhaustively searching the space to identify the global free energy minimum, as the number of sequences grows exponentially as $4^n$. 
 
\begin{figure}
  \centering
    \includegraphics[width=.425\textwidth]{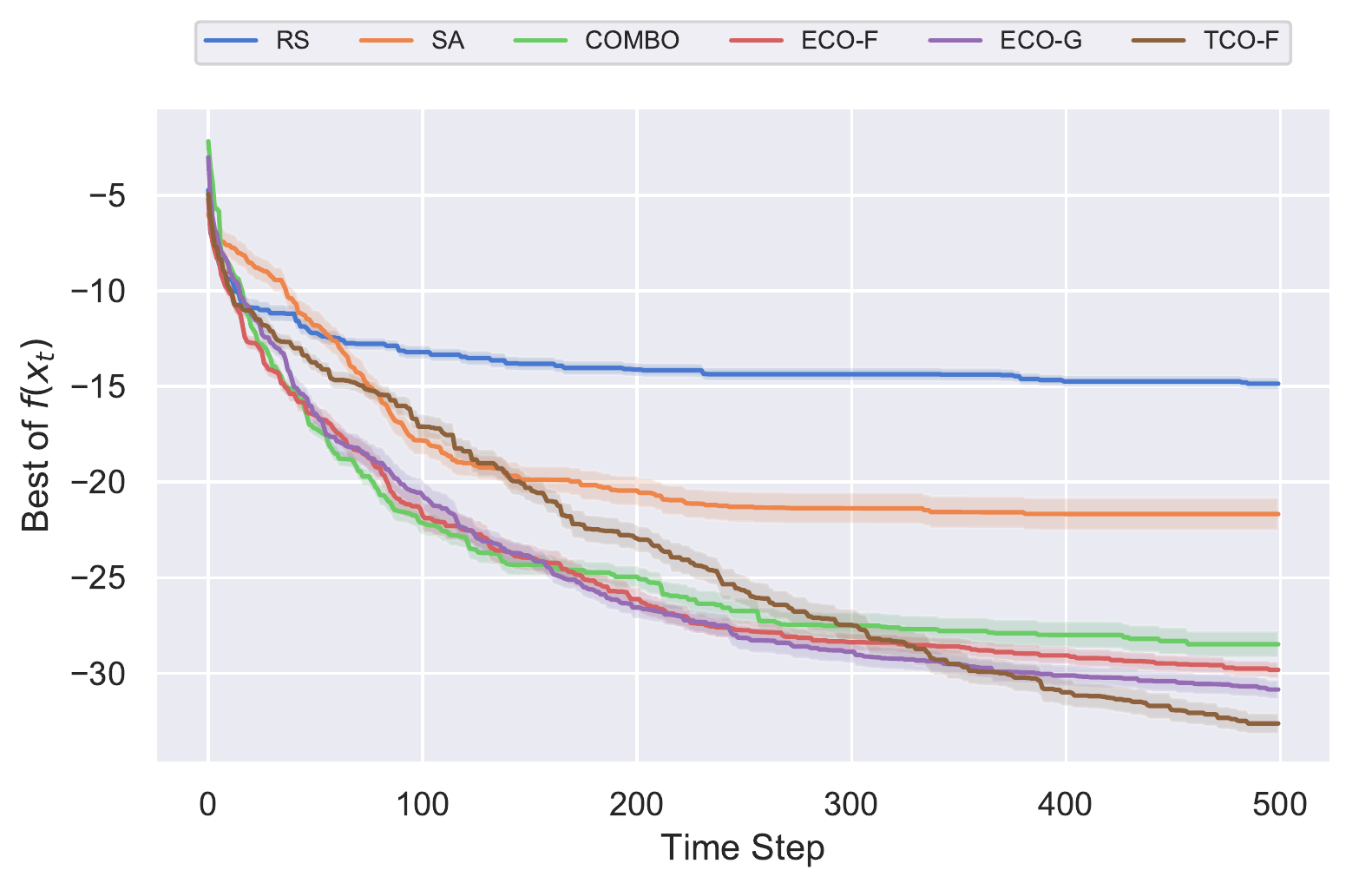}
    \caption{{\small RNA BBO Problem with $n = 30$}}
    \label{fig:sequence}
 \vspace{-2mm}
 \end{figure}

We formulate the RNA sequence optimization problem as follows: For a sequence of length $n$, find the RNA sequence which folds into a secondary structure with the lowest MFE.
In our experiments, we set $n = 30$ and $k = 4$. 
We then use the popular RNAfold package \cite{ViennaRNA} to evaluate the MFE for a given sequence. The goal is to find the lowest MFE sequence by calling the MFE evaluator minimum number of times.  
As shown in Figure \ref{fig:sequence}, both ECO-F and particularly ECO-G outperform the baselines as well as COMBO by a large margin. At higher number of evaluations, TCO-F beats the rest of the algorithms, which can be attributed to its exploration-exploitation trade-off.
See App. \ref{sec:BOCS_comparison} for comparison between TCO-F and BOCS+vanilla one-hot encoding.

\textbf{RNA Design Experiments}:
The problem is to find a primary RNA sequence $\phi$ which folds into a target structure $\omega$, given a folding algorithm $F$. Such target structures can be represented as a sequence of dots (for unpaired bases) and brackets (for paired bases).
In our algorithm, the action sets are defined as follows. For unpaired sites $\mathcal{A}_t = \{A, G, C, U\}$ and for paired sites $\mathcal{A}_t = \{GC, CG, AU, UA\}$. At the beginning of each run of our algorithm (ECO-F/G, TCO-F in conjunction with MCTS), we draw a random permutation for the order of locations to be selected in each level of the tree. The reward value offered by the environment (i.e. the black-box function) at any time step $t$ corresponds to the normalized Hamming distance between the target structure $\omega$ and the structure $y_t = F(x_t)$ of the sequence $x_t$ found by each algorithm, i.e. $d_H(\omega, y_t)$.

We compare the performance of our algorithms against RS as a baseline, where random search is carried out over the given structure (i.e. default policy $\pi^{\texttt{RS}}$) rather than over unstructured random sequences.
We also include two state-of-the-art algorithms in our experiments: MCTS-RNA \cite{MCTSRNA} and LEARNA \cite{runge2018learning}. MCTS-RNA has an exploration parameter, which we tune in advance (per sequence). LEARNA has a set of $14$ hyper-parameters tuned a priori using training data and is provided in \cite{runge2018learning}. Note that the latter training phase (for LEARNA) as well as the former exploration parameter tuning (for MCTS-RNA) are offered to the respective algorithms as an advantage, whereas for our algorithm we use a global set of heuristic choices for the two hyper-parameters, rather than attempting to tune the two hyper-parameters. In particular, we set the exploration parameter $c$ to $0.5$ for ECO and $0.25$ for TCO; and the number of MCTS playouts at each time step to $30 \times h$, where $h$ is the tree height (i.e. number of dots and bracket pairs). The latter heuristic choice is made since the bigger the tree, the more playouts are needed to explore the space.

We point out that the entire design pipeline in state-of-the-art algorithms typically also includes a local improvement step (as a post-processing step), which is either a rule-based search (e.g. in \cite{MCTSRNA}) or an exhaustive search (e.g. in \cite{runge2018learning}) over the mismatched sites. We do not include the local improvement step in our experiments, since we are interested in measuring sample efficiency of different algorithms. In other words, the question is the following: given a fixed and finite evaluation budget, which algorithm is able to get closer to the target structure. 

\begin{figure}[ht]
\centering
\includegraphics[width=.9\linewidth]{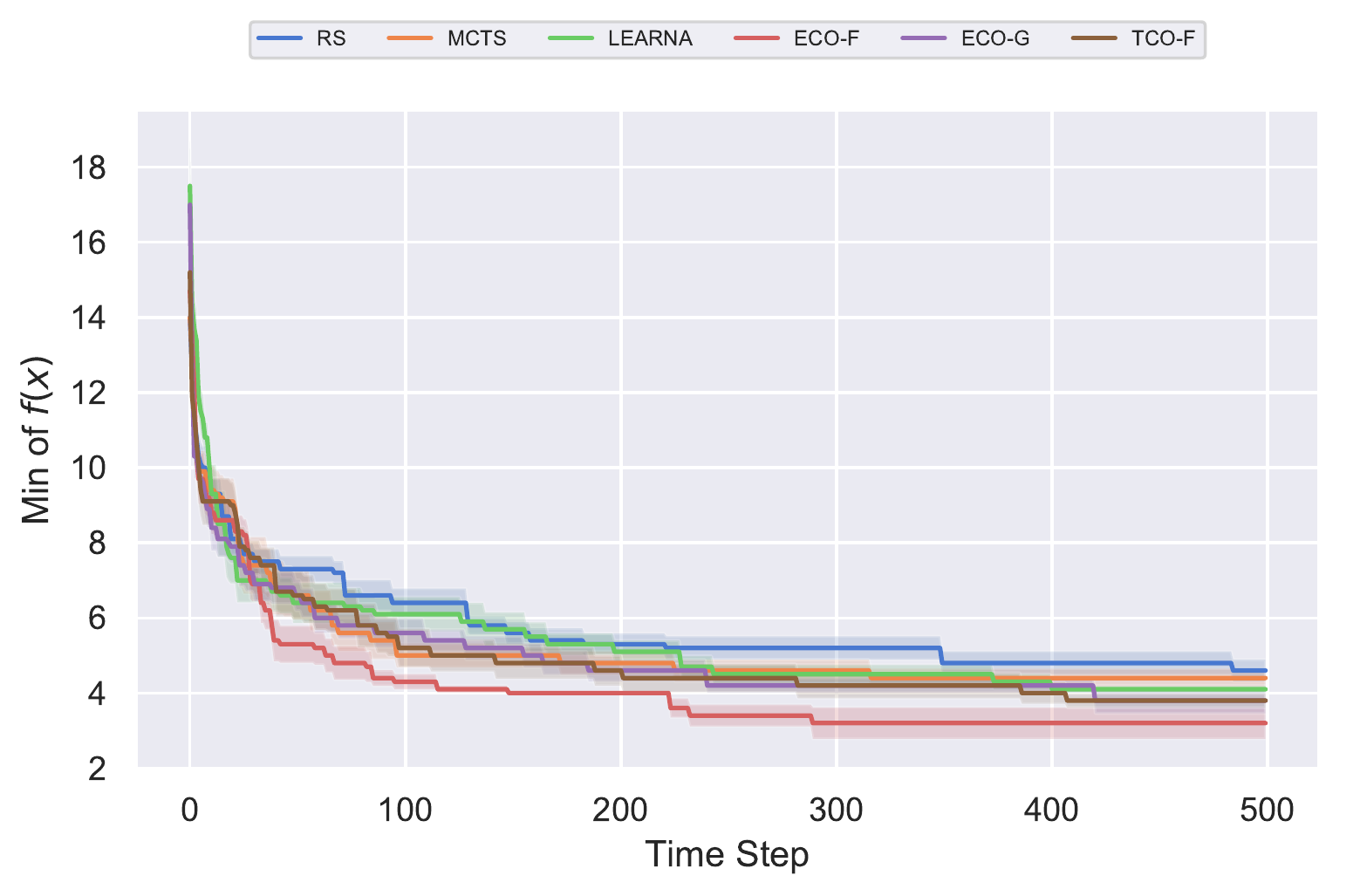}
\caption{Design puzzle $\#15$  with $n = 30$.}
\label{fig:design_15_main}
\vspace{-4mm}
\end{figure}
\begin{figure}[ht]
  \centering
    \includegraphics[width=.425\textwidth]{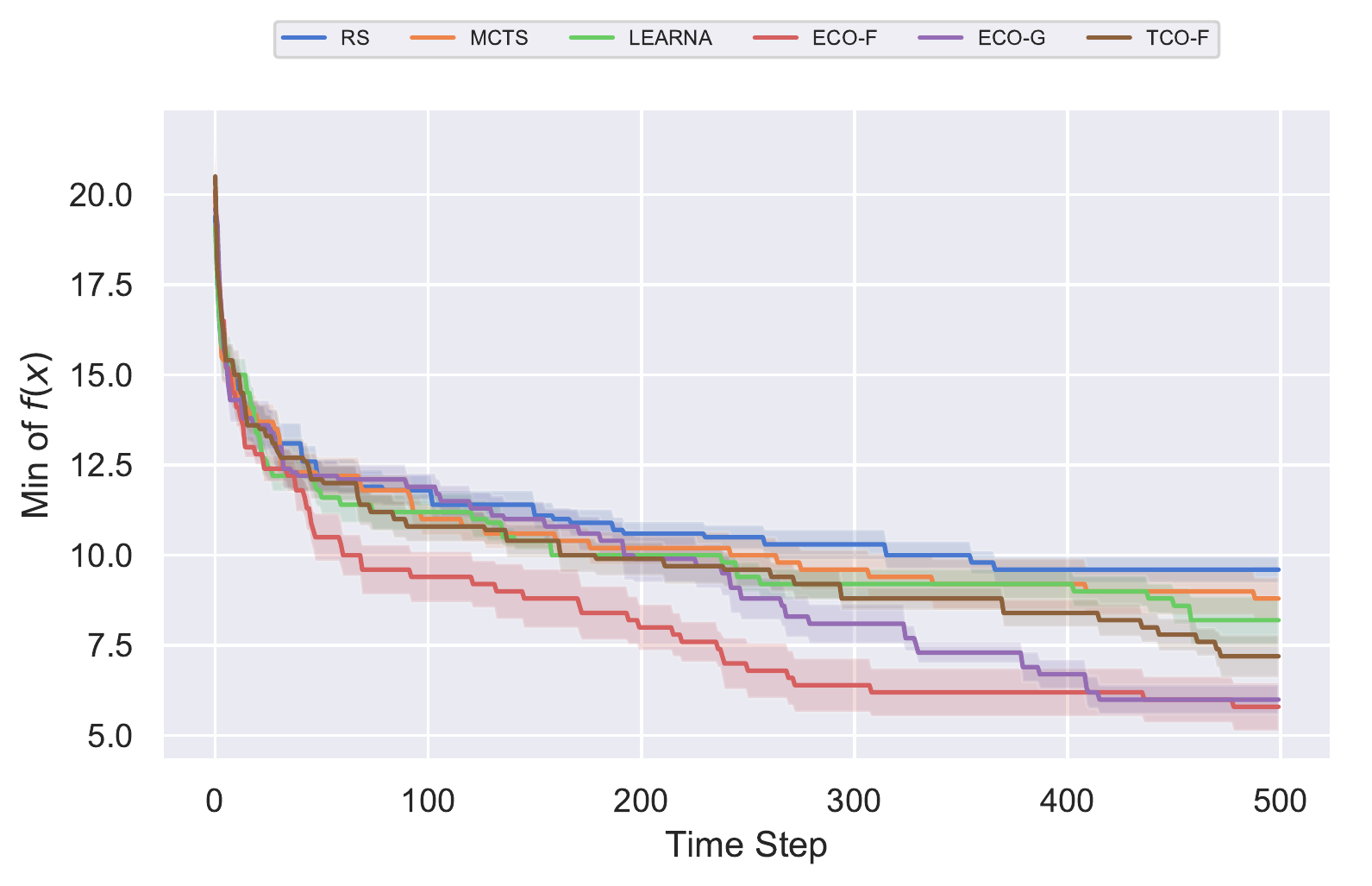}
    \caption{Design puzzle $\#41$ with $n = 35$.}
    \label{fig:design_41_main}
\end{figure}

In our experiments, we focus on three puzzles from the Eterna-100 dataset \cite{MODENA}. Two of the selected puzzles ($\#15$ and $\#41$ of lengths $30$ and $35$, resp.), despite their fairly small lengths, are challenging for many algorithms (see \cite{MODENA}). In both puzzles, ECO-F, ECO-G, and TCO-F (with MCTS as AFO) are able to significantly improve the performance of MCTS when limited number of black-box evaluations is available. All algorithms outperform RS as expected. Within the given $500$ evaluation budget, TCO-F, ECO-G, and esp. ECO-F, are superior to LEARNA by a substantial margin (see Figure \ref{fig:design_15_main}). In puzzle $41$ (Figure \ref{fig:design_41_main}), again both ECO-G and ECO-F significantly outperform LEARNA over the given number of evaluations. ECO-F is able to outperform LEARNA throughout the evaluation process, and on average finds a far better final solution than LEARNA. With MCTS as AFO, ECO algorithms outperform TCO, which can be attributed to the exploratory behavior of the latter in both AF (via TS) and AFO (via MCTS). See Appendix for puzzle $\#70$.

\section{Conclusions and Future Work}

We propose two novel Fourier representations as surrogate models for black box optimization over categorical variables and show performance improvements over existing baselines when combined with 
search algorithms. We utilize two algorithms to learn such surrogate models. Our ECO algorithm incorporates a computationally-efficient online estimator with strong adversarial guarantees (see \cite{COMEX} for theoretical results in the Boolean case), which can be shown to carry over to our setting as well. 
Our TCO algorithm uses a Bayesian regression model with a regularized horseshoe prior and selects queries via Thompson sampling to balance exploration-exploitation trade-off in surrogate model learning at the price of a higher computation cost.


Considering the performance variability with respect to different algorithms and representations across different problems, an important research avenue would be to derive an ensemble of such models rather than a single one and investigate potential performance gains. Such an ensemble model would then update and explore all models simultaneously and draw samples from either individual or a combination of models at any given time step. 

\clearpage
\appendix


\section{Proofs}
\label{app:proofs}

\textbf{Proof of Theorem 3.1}: We first assume that the one-hot variables $x_{ij} \in \{0, 1\}$. Plugging different choices of $x \in \mathcal{X} = [k]^n$ into Equation \eqref{eq:factorization} leads to a system of linear equations with $k^n$ unknowns (coefficients $\alpha_{\mathcal{I}, \mathcal{J}}$) and $k^n$ equations. We can express this system in matrix form as the product of the matrix of monomials (where each column $j$ corresponds to a monomial $\psi_j$, and each element $(i, j)$ corresponds to the evaluation of the monomial $\psi_j$ at the $i$-th choice for $x$) and the vector of unknown coefficients, which is set equal to the function values at the corresponding choices for $x$. We claim that there exists a permutation of the rows and columns of the matrix of monomials such that the latter becomes unit lower triangular, and is thereby full rank. As a result, the representation in \eqref{eq:factorization} for $x_{ij} \in \{0, 1\}$ is complete and unique.

To formally show that this permutation for the monomials' matrix exists, we use a construction by induction over the number of variables $n$ included in the representation. We denote the monomials' matrix over $\ell$ variables with $\Phi_{\ell}$, and define the one-hot variables $x_{ij} (j \in [k-1]$) as the \textit{descendants} of the \textit{parent} categorical variable $x_i$ ($\forall i \in [n]$). It is easy to see that such a construction for the base case of only one variable exits, as the monomials' matrix is a $k \times k$ matrix. In this case, we use the following permutation of the monomials (columns): $(1, x_{11}, x_{12}, \ldots, x_{1(k-1)})$. We also use the following permutation of the $x_1$ values in rows: $(k, 1, \dots, k-1)$. As a result, in this matrix only the elements of the first column and the main diagonal are non-zero and equal to one, and thus the matrix $\Phi_1$ is unit lower triangular.

Assuming that the induction hypothesis holds for $n$ variables, we show that it also holds for $n+1$ variables. Starting from a unit lower triangular matrix $\Phi_n$, we can construct the matrix $\Phi_{n+1}$ as follows. Note that all the $k^n$ columns of $\Phi_n$ correspond to the monomials composed of the descendants of the first $n$ variables $\{x_{ij}: \forall i \in [n] \; \textrm{and} \; \forall j \in [k-1] \}$, whereas each of the additional $k^{n+1} - k^n$ columns introduced in $\Phi_{n+1}$ involves exactly one term from $\{x_{(n+1)j}: \forall j \in [k-1]\}$ (possibly also containing factors from the previous $n$ variables). We can express $k^{n+1} - k^n$ using the following binomial expansion:
\begin{equation}
    k^{n+1} - k^n = (k-1) \sum_{m=1}^{n+1} \binom{n}{m-1} (k-1)^{m-1}.
\label{eq:binomial}
\end{equation}
In words, the additional $k^{n+1} - k^n$ columns can be considered as a collection of $\binom{n}{m-1} (k-1)^{m-1}$ $m$-th order monomials ($m \in [n+1]$), each of which includes one out of the $k-1$ descendant variables $x_{(n+1)j}$ ($j \in [k-1]$) together with $m-1$ variables from the (descendants) of the previous $n$ variables. Each of the latter $m-1$ variables can take values in $[k-1]$, whereas the remaining $n-m+1$ variables are set to $k$. 

Starting with $m=1$, we have $(k-1)$ first order terms $(x_{(n+1)1}, x_{(n+1)2}, \ldots, x_{(n+1)(k-1)})$ which we assign to columns $(k^n+1, \ldots, k^n+(k-1))$. The $x$ values associated with rows $(k^n+1, \ldots, k^n+(k-1))$ are constructed by assuming that $(i)$ $x_{n+1}$ takes values from $1$ to $k-1$ (in order), while $(ii)$ all the remaining $x_j$ ($j \in [n]$) variables are set to $k$. As a consequence of $(i)$, all the elements on the main diagonal are ones; also, as a consequence of $(ii)$, all the higher degree monomials involving $x_{n+1}$ (which occupy the elements after the diagonal ones) are equal to zeros. Thus the augmented matrix, until this point, remains unit lower triangular. 

We then consider the second degree terms $m=2$, where we have $n (k-1)$ terms (choice of one out of the other $n$ variables, each taking values from $[k-1]$) for each of the $k-1$ choices for the variable $x_{n+1}$. Starting with second order monomials involving $x_1$ and $x_{n+1}$, we again assume that all the remaining variables $x_i$ ($i \not\in \{1, n+1\} $) are equal\footnote{Note that this assumption is necessary in order to ensure that monomials in future columns for the same row are evaluated to zero; choices of $x$ where this assumption is not valid is addressed in next rows.} to $k$. For any choice of $(x_1, x_{n+1}) \in [k-1]^2$, we add a new column corresponding to the monomial $x_{1j} x_{(n+1)\ell}$ as well as a new row in which $x_i = j$ and $x_{n+1} = \ell$, whereas the remaining variables are set to $k$. As a result, we have that: $(i)$ the diagonal element in the new row/column is equal to one, and $(ii)$ all the elements in the future\footnote{Note that the elements in the previous columns in the same row corresponding to monomials involving the selected $m-1$ variables are non-zeros as well.} columns are zeros, since any combination of one of the descendants of $x_{n+1}$ with the remaining variables (any variable except $x_1$ and $x_{n+1}$) is zero. We continue this construction strategy for the remaining variables until all the second degree terms involving $x_{n+1}$ and one out of the remaining $n-1$ variables is exhausted. We then repeat the same idea for terms with orders up to $n+1$, as defined by the binomial expansion in $\eqref{eq:binomial}$. As a result of this construction strategy, the monomial matrix $\Phi_{n+1}$ is unit lower triangular.



Now, we use this result to show the completeness and uniqueness of the representation with one-hot variables in $\{-1, 1\}$ in the following way.

\textbf{Completeness}: We showed that the representation \eqref{eq:factorization} over $\{0, 1\}$ is complete, i.e. we can express any function using the representation in \eqref{eq:factorization}, where we have at most one descendant term from the same parent in each monomial. Now, we replace each $x_{ij}$ (from $\{0, 1\}$) in the latter representation with $\nicefrac{(1 - x^{\prime}_{ij})}{2}$, where $x^{\prime}_{ij} \in \{1, -1\}$. The new representation can also be expressed via the expansion \eqref{eq:factorization} since no two descendants from the same parent variable are being multiplied with each another.

\textbf{Uniqueness}: Assume that the uniqueness condition is not satisfied. Then we have two distinct polynomial representations $f_1$ and $f_2$ that have the same value for every $x$. However, since $f_1$ and $f_2$ are distinct polynomials, $f_1(x) - f_2(x)$ is a polynomial $p(x)$ which is non-zero in at least one input $x^*$. This implies that $f_1(x^*) - f_2(x^*)$ is also non-zero, which is a contradiction, and the proof is complete.

\begin{remark} 
The group-theoretic Fourier representation defined in Equation \eqref{eq:group_fourier_complex} is unique and complete. 
\end{remark}
\begin{proof}
Let $\chi = [k]^n$ be the categorical domain. Let the true function be $f$. For generality, let us consider a complex valued function $f: \chi \rightarrow \mathbb{C}$ where $\mathbb{C}$ is the field of complex numbers. The basis functions are $\psi_{{\cal I}} (x)$. Now, one can view a function as a $[k]^n$-length vector,
one entry each for evaluating the function at every point in the domain $\chi$. We denote the vector for function $f$, thus obtained, by $f^{\chi} \in \mathbb{C}^{k^{n}}$. Similarly, denote the vector for evaluations of the basis function $\psi_{{\cal I}}$ by $\psi_{{\cal I}}^{\chi} \in \mathbb{C}^{k^{n}}$. Let $A$ be a matrix created by stacking all vectors corresponding to basis vectors in the columns. Then, the Fourier representation can be written as $f^{\chi} = A \alpha$ where $\alpha$ is the vector of Fourier coefficients in our group-theoretic representation. Now, due to the use of complex exponentials, one can show that $\sum_{x \in [k]^n} \psi_{{\cal I}} (x) \psi_{{\cal I}'} (x) = 0$ if ${\cal I} \neq {\cal I}'$. Therefore, the columns of the matrix $A$ are orthogonal. Hence, $A$ is a full rank matrix. Therefore, our representation is merely representing a vector in another full rank orthogonal basis. Hence, it is unique and complete.
\end{proof}

\section{Description of Algorithms}
\label{app:descriptions}

\textbf{Surrogate Model Learning Algorithm via Online Regression}:
Let $n$ and $k$ denote the number of variables and the cardinality of each variable, respectively. The surrogate models used in ECO-F and ECO-G correspond to approximations of the representations given in \eqref{eq:factorization} and \eqref{eq:group_fourier}, respectively, where each approximation is obtained by restricting the maximum order of interactions among variables to $m$. We consider each term in the latter surrogate models, i.e. monomials $\psi_{\mathcal{I}, \mathcal{J}}$ from \eqref{eq:monomials} in ECO-F and characters $\psi_{\beta, \mathcal{I}} (\beta \in \{r, i\})$ from \eqref{eq:characters} in ECO-G, as an expert, denoted by $\psi_i$ ($i \in [d]$).
The number of such experts in ECO-F is $d = \sum_{i=0}^m \binom{n}{i} (k-1)^i$ which coincides with the dimensionality of the space $k^n$ when $m = n$, whereas the number of experts in ECO-G is equal to $d = 2 \sum_{i=0}^m \binom{n}{i} (k-1)^i - 1$. 
The coefficient of each expert $\psi_i$ is designated by $\alpha_i$. Since the exponential weights, utilized to update the coefficients $\alpha_i$, are non-negative, we maintain two non-negative coefficients $\alpha_i^{+}$ and $\alpha_i^{-}$, which yield $\alpha_i = \alpha_i^{+} - \alpha_i^{-}$.

We initialize all the coefficients with a uniform prior, i.e. $\alpha_i^{\gamma} = \nicefrac{1}{2d}$ ($\forall i \in [d] \, \textrm{and} \, \gamma \in \{-, +\}$).
In each time step $t$, we draw a sample $x_t$ via Algorithm \ref{algo:SA} with respect to our current estimate for the surrogate model $\widehat{f}_{\alpha}$. The latter sample is then plugged into the black-box function to obtain the evaluation $f(x_t)$. This leads to a mixture loss $\ell^t$ as the difference between the evaluations obtained by our surrogate model and the black-box function for query $x_t$. Using this mixture loss, we compute the individual loss $\ell_i^t$ for each expert $\psi_i$. Finally, we update each coefficient in the model via an exponential weight obtained according to its incurred individual loss. We repeat this process until stopping criteria are met.

\begin{algorithm}
\caption{Expert Categorical Optimization}
\begin{algorithmic}[1]
\STATE \textbf{Inputs:} sparsity $\lambda$, max model order $m$
\STATE $t \gets 0, \: \forall \gamma \in \{-, +\} \: \, \forall i \in [d]: \alpha^t_{i, \gamma} \gets \tfrac{1}{2 d}$
\REPEAT
\STATE $x_t \sim \widehat{f}_{\alpha^t} \;$ via Algorithm \ref{algo:SA} or Algorithm \ref{algo:MCTS}
\STATE Observe $\, f(x_t)$
\STATE $\widehat{f}_{\alpha^t}(x) \gets \sum_{i \in [d]} \big (\alpha^{t}_{i, +} - \alpha^{t}_{i, -} \big )~\psi_i(x) $
\STATE $\ell^{t+1} \gets \widehat{f}_{\alpha^t}(x_t) - f(x_t)$
\FOR {$i \in [d] \: \textrm{and} \: \gamma \in \{-, +\}$}
\STATE $\ell_i^{t+1} \gets 2 \, \lambda \, \ell^{t+1} \, \psi_i(x_t)$
\STATE $\alpha^{t+1}_{i, \gamma} \gets \alpha^{t}_{i, \gamma} \exp \big (- \,\gamma \, \eta_t \, \ell_i^{t+1} \big)$
\STATE $\alpha^{t+1}_{i, \gamma} \gets \lambda \cdot \tfrac{\alpha^{t+1}_{i, \gamma}}{\sum_{\mu \in \{-, +\}} \sum_{j \in [d]} \alpha^{t+1}_{j, \mu}}$
\ENDFOR
\STATE $t \gets t + 1$
\UNTIL{Stopping Criteria}
\RETURN $\widehat{x}* = \arg\min_{\{x_i : \, \forall i \in [t]\}} f(x_i)$
\end{algorithmic}
\label{algo:ECO}
\end{algorithm}

\textbf{Number of Characters in Representations}: 
The number of terms in vanilla one-hot encoded Fourier representation is $2^{kn}$, whereas our abridged representation reduces this number to $k^n$ matching the space dimensionality, thereby making the algorithms computationally tractable and efficient. 
When a max degree of $m$ is used in the approximate representation, the number of terms in the abridged representation is equal to $d = \sum_{i=0}^m \binom{n}{i} (k-1)^i$. The corresponding number in a vanilla one-hot encoded representation is equal to $\sum_{i=0}^m \binom{nk}{i}$. Finally, the numbers of terms in the full and order-$m$ group-theoretic Fourier expansions are equal to $2 k^n - 1$ and $d = 2 \sum_{i=0}^m \binom{n}{i} (k-1)^i - 1$, respectively.

\textbf{Surrogate Model Learning Algorithm via Sparse Bayesian Regression}: We use our surrogate model $ f(x)=\sum_{i=1}^d \alpha_i \psi_i(x)$ to approximate the black-box function, where $\psi_i$ are the characters of the representation, $d$ is the number of characters in the representation, and $\alpha_i$ are the coefficients. We utilize the abridged one-hot encoded Fourier representation for this model due to its smaller number of characters which makes it much more scalable. We assume that one-hot encoded variables take values in $\{0, 1\}$. Note that from the proof of the abridged one-hot encoded Fourier representation in Appendix \ref{app:proofs}, both assignments $\{1, -1\}$ and $\{0, 1\}$ for one-hot encoded variables are permissible in the latter representation.

We assume that the approximation errors between the black-box evaluations $y$ at points $x$ and our surrogate function values $f(x)$ are independent and identically distributed as follows:
\begin{equation}
    y - f(x) \sim \mathcal{N}(0, \sigma). 
\end{equation}
We estimate the coefficients of our surrogate approximation using Bayesian linear regression over the black-box evaluation data seen so far. To avoid high-variance estimators for the coefficients due to data scarcity, sparsity-inducing priors are recommended in \cite{BOCS}. We use the regularized horseshoe model \cite{finnish_horseshoe} defined as follows:
\begin{eqnarray*}
    \alpha_i | \lambda_i, \tau, c &\sim& \mathcal{N}(0, \tau^2 \tilde{\lambda_i}^2) \quad ~~(\forall j \in [d])\\ 
    \tilde{\lambda}_i^2 &=& \tfrac{c^2 \lambda_i^2}{c^2 + \tau^2 \lambda_i^2} \\
    \lambda_i &\sim& \mathcal{C}^+(0, 1) \\
    c^2 &\sim& \mathcal{IG}(\nicefrac{\nu}{2}, \nicefrac{\nu s^2}{2}) \\
    \tau &\sim& \mathcal{C}^+(0, \tau_0) \\
    \tau_0 &=& \tfrac{d_0}{d - d_0} \tfrac{\sigma}{\sqrt{t}} \\
    \sigma &\sim& \mathit{Exp}(1)
\end{eqnarray*}
where $\mathcal{C}^+(0, 1)$ is the standard half-Cauchy distribution, and $t$ is the number of observations seen so far. In this model, the global $\tau$ and the local $\lambda_i$ hyper-parameters individually shrink the magnitude of regression coefficients $\alpha_i$. The hyper-parameter $d_0$ designates the expected number of relevant coefficients, but since the sparsity induced by the corresponding threshold is only weakly sensitive to $d_0$, a rough estimate is typically sufficient. In our experiments, we set $\nu = s = 1$, and $d_0 =  d/100$.

Given the above Bayesian model, the TCO-F algorithm works as follows. At each time step $t$, we draw a sample for the coefficients $\alpha_i$ ($i \in [d]$) from the posterior. We use the No-U-Turn Sampler (NUTS) \cite{NUTS} in order to sample the coefficients efficiently. Given the latter sample, we then construct the approximation function instance $f(x) = \sum_{i=1}^d \alpha_i \psi_i(x)$. Following \cite{BOCS}, we then select our next point for black-box evaluation by minimizing a regularized version of this approximation function, i.e. $f(x) + \lambda \|x\|_1$, where $\|x\|_1$ designates the $\ell_1$-norm of $x$, and $\lambda$ is a regularization hyperparameter. Throughout our experiments, we set $\lambda = 10^{-5}$. The AFO in either algorithm \ref{algo:SA} or algorithm \ref{algo:MCTS} is then used to carry out this minimization for the generic BBO or the design problem, respectively.

\textbf{MCTS Algorithm}: For a complete version of Algorithm \ref{algo:MCTS}, see Algorithm \ref{algo:MCTS_full}.


\begin{algorithm}
\caption{MCTS with Surrogate Reward Model}
\begin{algorithmic}[1]
\STATE \textbf{Inputs:} surrogate reward model $\widehat{f}_{\alpha}$, exploration parameter $c$, search tree $\mathcal{T} $
\STATE $s_{\texttt{best}}=\{\}$,  $r_{\texttt{best}}=-\infty$ 
\REPEAT  
\STATE Initialize episode $t = 0$, $s_t = []$
\WHILE{ $s_t \notin \mathcal{T}$ }  
\STATE $a_t \gets \pi^\mathcal{T}(s_t) =   
\arg\max_{a \in \mathcal{A}_t} Q(s_t, a) + c \sqrt{\nicefrac{\ln{N(s_t)}}{N(s_t, a)}}$
\STATE $s_{t+1} \gets T(s_t, a_t) = s_t \circ a_t $
\STATE $t \gets t+1$
\ENDWHILE
\STATE $s_{\texttt{leaf}} = s_{t}$ 
\STATE $\mathcal{T} \gets \mathcal{T} \cup \{s_{t}\}$
\STATE $\forall a \in \mathcal{A}_{t+1}: N(s_{t}, a) = 0 $,\, $Q(s_{t}, a) =0$
\REPEAT  
\STATE $a_t \gets \pi^{RS}(s_t)$
\STATE $s_{t+1} \gets T(s_t, a_t) = s_t \circ a_t $
\STATE $t \gets t+1$
\UNTIL{ $s_t$ is terminal }
\STATE $r \gets - \widehat{f}_{\alpha}(s_t)$
\STATE $s \gets s_{\texttt{leaf}}$
\REPEAT
\STATE $N(s, a) \gets N(s, a) + 1 $
\STATE $Q(s, a) \gets Q(s, a) + \frac{1}{N(s,a)}(r - Q(s, a))$
\STATE $s \gets \texttt{parent}(s)$; $a \gets$  visited action on $s$
\UNTIL{$s$ is the root node}
\IF{$r > r_{\texttt{best}}$} 
\STATE $r_{\texttt{best}} \gets r \: \texttt{and} \:  s_{\texttt{best}} \gets s_t$
\ENDIF
\UNTIL{Stopping Criteria}
\RETURN $s_{\texttt{best}}$
\end{algorithmic}
\label{algo:MCTS_full}
\end{algorithm}


\textbf{Learning Rate in ECO}: The anytime learning rate (at time step $t$) used in Algorithm \ref{algo:ECO} is given by \cite{adaptive_EG, COMEX}:
\begin{equation}
	\eta_t = \min \bigg \{ \frac{1}{e_{t-1}}, c \sqrt{\frac{\ln{(2 \, d)}}{v_{t-1}}} \bigg \},
\end{equation}
where $c \overset{\Delta}{=} \sqrt{2(\sqrt{2} - 1)/(\exp(1)-2)}$ and 
\begin{align*}
    z_{j, t}^{\gamma} &\overset{\Delta}{=} - 2 \, \gamma \, \lambda \, \ell_t \, \psi_j(x_t) \\
    e_t &\overset{\Delta}{=} \inf_{k \in \mathbb{Z}} \bigg \{ 2^k: 2^k \geq \max_{s \in [t]} \max_{ \substack{j, k \in [d] \\ \gamma, \mu \in \{-, +\}}} | z_{j, s}^{\gamma} - z_{k, s}^{\mu} |  \bigg \} \\
    v_t &\overset{\Delta}{=}  \sum_{s \in [t]} \sum_{\substack{j \in [d] \\ \gamma \in \{-, +\}} } \alpha_{j, s}^{\gamma} \bigg ( z_{j, s}^{\gamma} - \sum_{\substack{k \in [d] \\ \mu \in \{-, +\}}} \alpha_{k, s}^{\mu} z_{k, s}^{\mu} \bigg )^2.
\end{align*}

\section{BBO Experiments}
\label{app:bbo_exp}

\textbf{Latin Square Problem}: A latin square of order $k$ is a $k \times k$ matrix of elements $x_{ij} \in [k]$, such that each number appears in each row and column exactly once. When $k = 5$, the problem of finding a latin square has $161,280$ solutions in a space of dimensionality $5^{25}$. We formulate the problem of finding a latin square of order $k$ as a black-box function by imposing an additive penalty of one for any repetition of numbers in any row or column. As a result, function evaluations are in the range $[0, 2 k (k-1)]$, and a function evaluation of zero corresponds to a latin square of order $k$. We consider a noisy version of this problem, where an additive Gaussian noise with zero mean and standard deviation of $\sigma = 0.1$ is added to function evaluations observed by each algorithm. 

\begin{figure}[ht]
\center
\includegraphics[width=.9\linewidth]{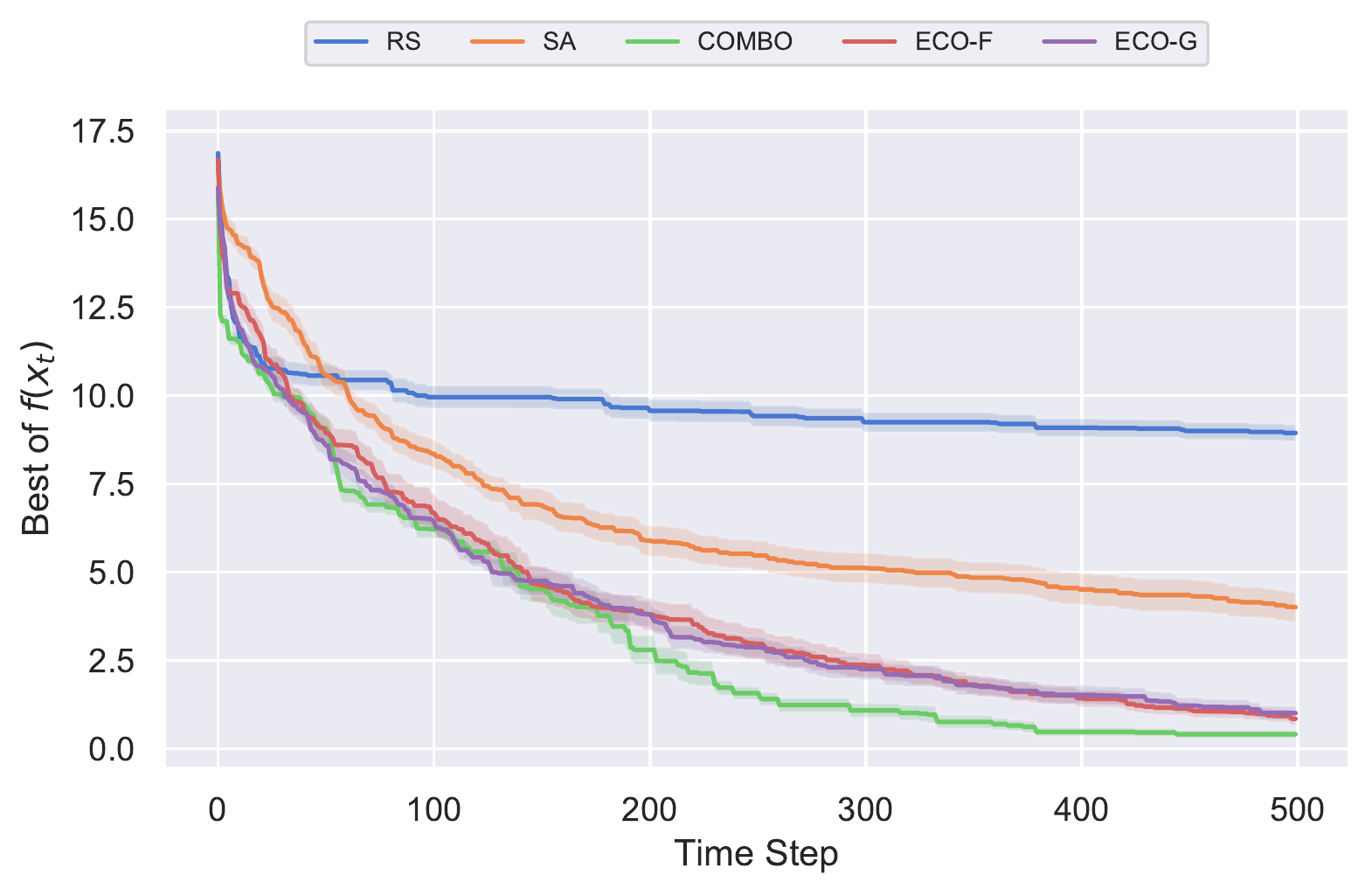} 
\caption{\small \label{fig:latin_square}
Best function evaluation seen so far for the Latin Square problem. Each time step $t$ corresponds to a black-box evaluation.}
\end{figure}

Figure \ref{fig:latin_square} demonstrates the performance of different algorithms, in terms of the best function value found until time $t$, over $500$ time steps. Both ECO-F and ECO-G outperform the baselines with a considerable margin. In addition, ECO-G outperforms COMBO at $130$ samples. At larger time steps, COMBO outperforms the other algorithms, however, this performance comes at the price of a far larger computation time. As demonstrated in Table \ref{table:times}, ECO-F and ECO-G offer a speed-up over COMBO by a factor of approximately $100$ and $50$, respectively. We note that TCO-F (not included in the plot) performs poorly (similar to RS) on this problem, which can be attributed to the strong promotion of sparsity by the regularized horseshoe prior and the fact that the Latin Square problem has a dense representation (we observed a similar behavior from the horseshoe prior of \cite{BOCS}).

\textbf{Pest Control Problem}: In the pest control problem, given $n$ stations and $k-1$ pesticide types, the idea is to maintain the spread of pest (with minimum cost), which is propagating throughout the stations in an interactive and probabilistic fashion. The $k$-th category for each variable corresponds to the choice of no pesticide at all. Controlling the spread of the pest is carried out via the choice of the right type of pesticide subject to a penalty proportional to its associated cost. A closed form definition of this problem is given in \cite{COMBO}. 

\begin{figure}[ht]
\center
\includegraphics[width=.9\linewidth]{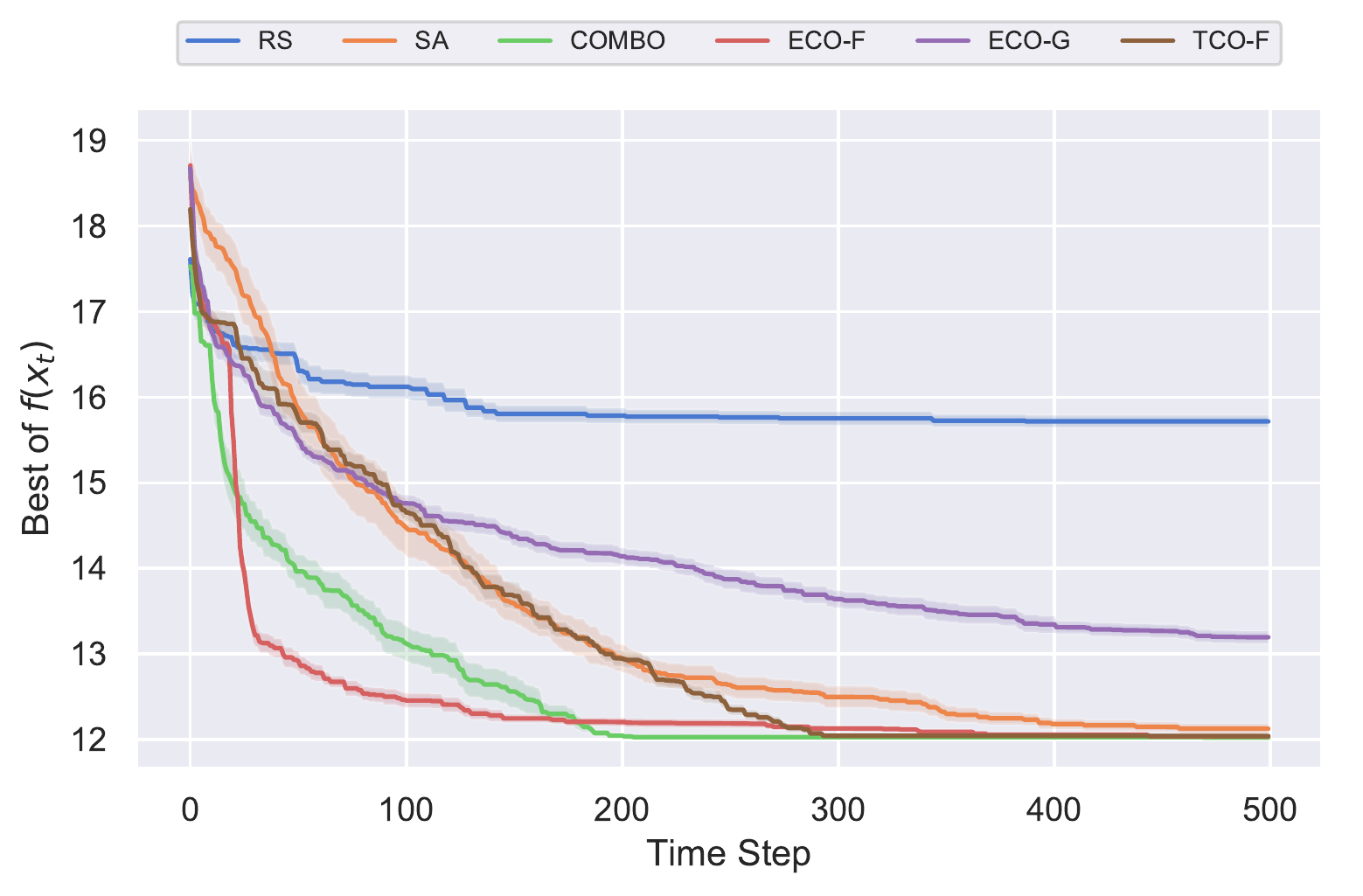}
\caption{\small \label{fig:pest_control}
Best function evaluation seen so far for the pest control problem.}
\end{figure}

The results for different algorithms are shown in Figure \ref{fig:pest_control}. Despite the fact that COMBO is able to find the minimum in fewer time steps (in $\approx 200$ steps) than ECO-F (in $\approx 360$ steps) on average, ECO-F outperforms COMBO during initial time steps (until $ t\approx180$). TCO-F is able to find the minimum faster than ECO-F and within $280$ steps. SA performs competitively, but eventually is unable to find the optimal solution to this problem over the designated $500$ steps. The poor performance of ECO-G can be explained by the interactive nature of the problem, where early mistakes are punished inordinately. Early mistakes made by ECO-G can also be attributed to the large number of experts (with noisy coefficients) in its model, which in turn promotes an early exploratory behavior.

\textbf{RNA Sequence Optimization Problem}: Structured RNA molecules play a critical role in many biological applications, ranging from control of gene expression to protein translation. 
The native secondary structure of a RNA molecule is usually the minimum free energy (MFE) structure. 
Consider an RNA sequence as a string $A=a_1 \ldots a_n$ of $n$ letters (nucleotides) over the alphabet $\Sigma=\{A, U, G, C\}$. A pair of complementary nucleotides $a_i$ and $a_j$, where $(i<j)$, can interact with each other and form a base pair (denoted by $(i, j)$), A-U, C-G and G-U being the energetically stable pairs.  Thus, the secondary structure of an RNA can be represented by an ensemble of pairing bases. 

 Finding the most stable RNA sequences has immediate applications in material and biomedical applications \cite{li2015rna}. Studies show that by controlling the structure and free energy of a RNA molecule, one may modulate its translation rate and half-life in a cell \cite{buchan2007halting, davis2008bioinformatic}, which is important in the context of viral RNA. A number of RNA folding algorithms \cite{ViennaRNA, markham2008unafold} use a thermodynamic model (e.g.  \cite{zuker1981optimal}) and dynamic programming to estimate MFE of a sequence. However, the  $O(n^3)$ time complexity of these algorithms  prohibits their use for evaluating substantial numbers of RNA sequences \cite{gould2014computational} and exhaustively searching the space to identify the global free energy minimum, as  the number of sequences grows exponentially as $4^n$. 
 
 Here, we formulate the RNA sequence optimization problem as follows:  For a sequence of length $n$, find the RNA sequence that will fold into the secondary structure with the lowest minimum free energy.
In our experiments, we initially set $n = 30$ and $k = 4$. We then use the popular RNAfold package \cite{ViennaRNA} to evaluate the MFE for a given sequence. The goal is to find the lowest MFE sequence by calling the MFE evaluator minimum number of times.  
The performance of different algorithms is depicted in Figure \ref{fig:sequence}, where both ECO-F and particularly ECO-G outperform the baselines as well as COMBO by a considerable margin. At higher number of evaluations, TCO-F beats the rest of the algorithms, which can be attributed to its exploration-exploitation trade-off.

\begin{figure}[ht]
\center
\includegraphics[width=.9\linewidth]{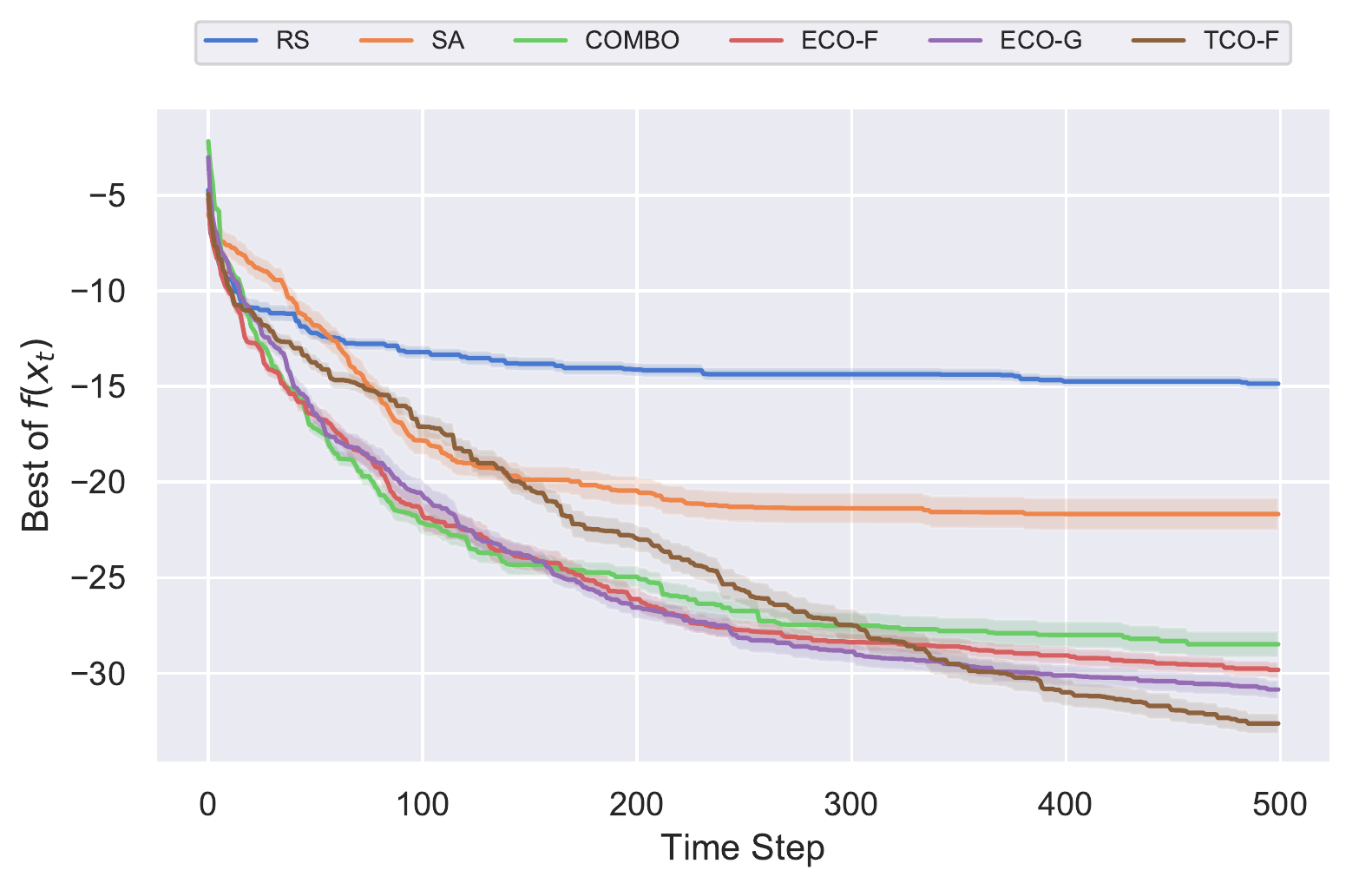}
\caption{\small \label{fig:RNA}
Best function evaluation seen so far for the RNA sequence optimization problem with $n = 30$.}
\end{figure}


\begin{table}[t]
\caption{Average computation time per step (in Seconds) over different problems and algorithms.}
\label{table:times}
\begin{center}
\begin{small}
\begin{sc}
\begin{tabular}{lccccccr}
\toprule
Data & Latin Square & Pest Cont. & RNA Opt. \\
\midrule
$n$ & 25 & 25 & 30 \\
$k$ & 5 & 5 & 4\\
COMBO & 170.4 & 151.0 & 253.8 \\
ECO-F & 1.5 & 1.4 & 2.0 \\
ECO-G & 3.6 & 3.3 & 5.7  \\
TCO-F & 55.7 & 53.2 & 67.0  \\
\bottomrule
\end{tabular}
\end{sc}
\end{small}
\end{center}
\end{table}

\begin{figure}[t]
  \centering
  \begin{minipage}[b]{0.45\textwidth}
    \hspace*{2cm}\includegraphics[width=.22\textwidth]{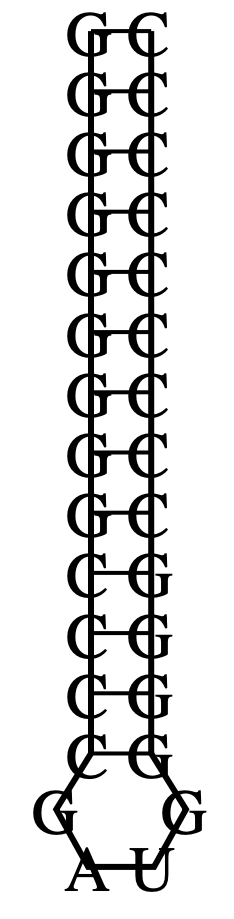}
    \caption{RNA Structure via ECO-G for $n = 30$}
    \label{fig:rna_30_fold}
  \end{minipage}
  \hfill
  \begin{minipage}[b]{0.45\textwidth}
    \hspace*{1.5cm}\includegraphics[width=.29\textwidth]{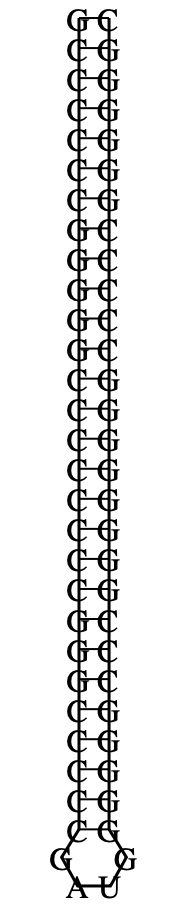}
    \caption{RNA Structure via ECO-G for $n = 60$}
    \label{fig:rna_60_fold}
  \end{minipage}
\end{figure}

\begin{figure}[t]
  \centering
  \begin{minipage}[b]{0.45\textwidth}
    \hspace*{1cm}\includegraphics[width=.7\textwidth]{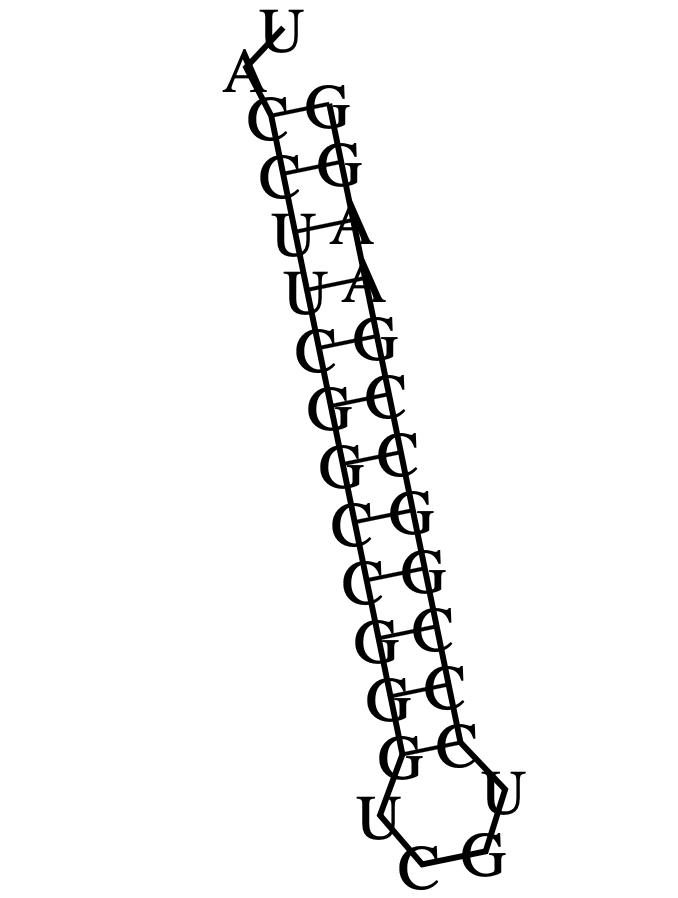}
    \caption{RNA Structure via SA for $n = 30$}
    \label{fig:rna_30_fold_sa}
  \end{minipage}
  \hfill
  \begin{minipage}[b]{0.45\textwidth}
    \hspace*{1.5cm}\includegraphics[width=.25\textwidth]{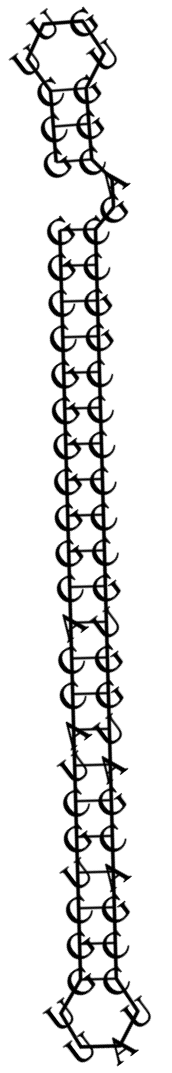}
    \caption{RNA Structure via SA for $n = 60$}
    \label{fig:rna_60_fold_sa}
  \end{minipage}
\end{figure}

\begin{figure}[t]
  \centering
  \begin{minipage}[b]{0.45\textwidth}
    \hspace*{2cm}\includegraphics[width=.43\textwidth]{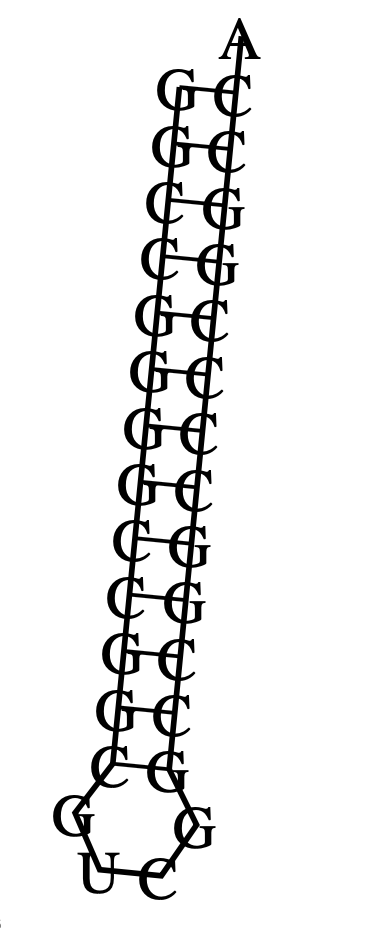}
    \caption{RNA Structure via ECO-G for $n = 31$}
    \label{fig:rna_31_fold}
  \end{minipage}
  \hfill
  \begin{minipage}[b]{0.45\textwidth}
    \hspace*{1.5cm}\includegraphics[width=.35\textwidth]{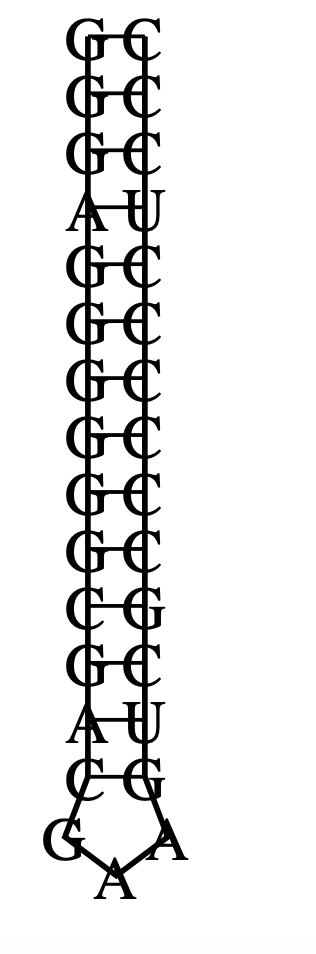}
    \caption{RNA Structure via ECO-G for $n = 31$}
    \label{fig:rna_31_fold_2}
  \end{minipage}
\end{figure}

\textbf{Energy-optimized RNA Structures}:
Sample RNA sequences obtained via ECO-G after $4000$ time steps for $n=30$ and $n=60$ are shown in Figures \ref{fig:rna_30_fold} and \ref{fig:rna_60_fold}, respectively. The resulting energy-optimized sequences (as obtained using RNAfold service) have high ($>90\%$) GC content that makes the strongest positive contribution to lowering MFE \cite{trotta2014normalization}, as pairings between G and C have three hydrogen bonds and are more stable compared to A and U pairings, which have only two. The final single-strand RNA sequence folds into a GC-paired double helix and a $4$ nucleotide long hairpin loop in the middle, which is a tetraloop reported in literature (\url{http://www.rna.icmb.utexas.edu/SIM/4C/energetics_new/}). Figures \ref{fig:rna_31_fold} and \ref{fig:rna_31_fold_2} show two sample structures of the ECO-G optimized sequences for $n=31$, again showing the same trend. For odd values of $n$, there is presence of a loop with an odd number of residues or a single unpaired base at the end, but there is still a GC-rich double helix. In contrast, the structures generated by the under-performing algorithms do show presence of unpaired bases and are less in GC content, leading to high energy structures (e.g. Figures \ref{fig:rna_30_fold_sa} and \ref{fig:rna_60_fold_sa} are obtained via SA after $4000$ steps for $n=30$ and $60$, respectively).

\section{Design Experiments}
\label{app:design_exp}

\begin{figure}[ht]
\center
\includegraphics[width=.9\linewidth]{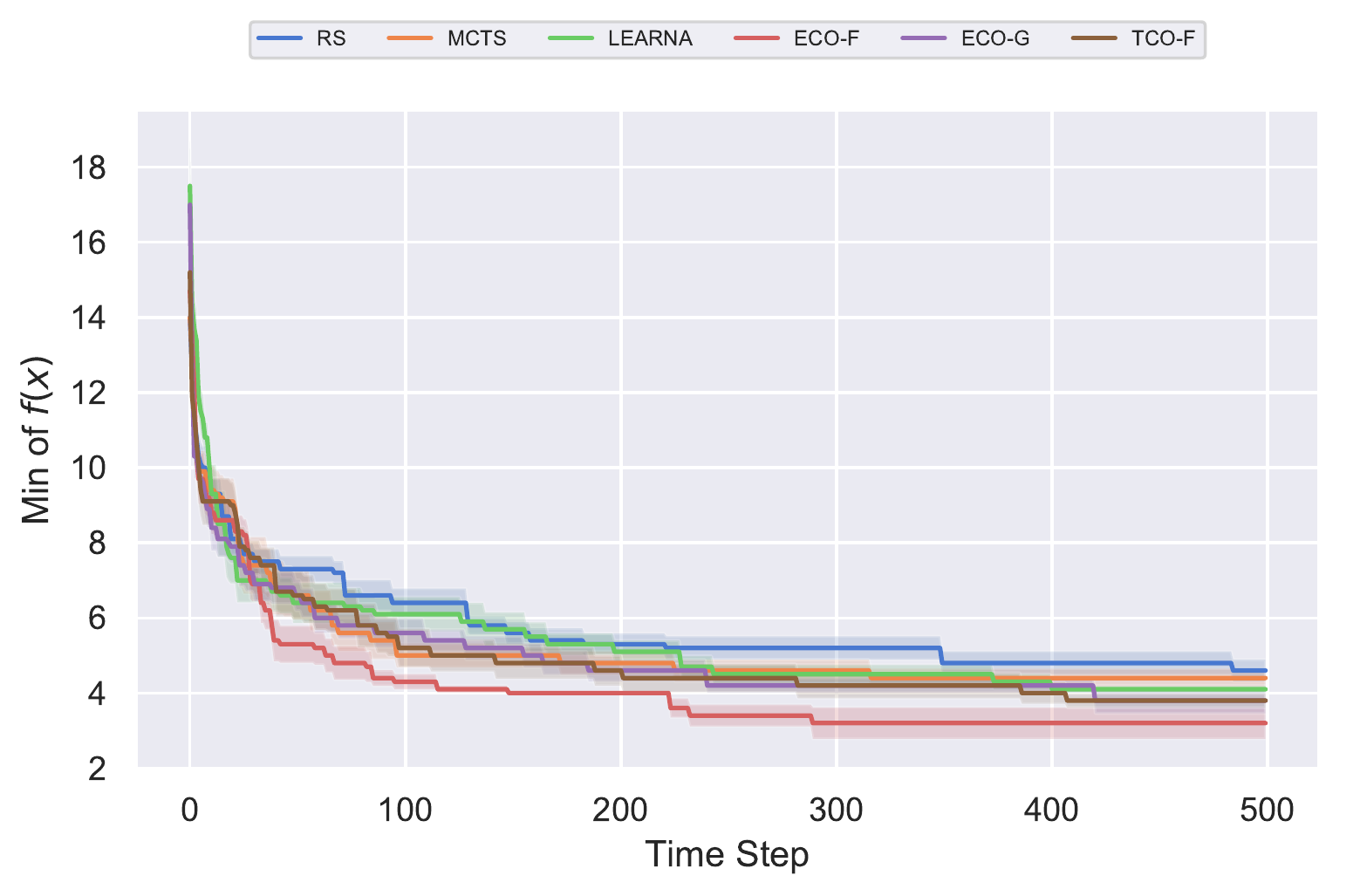}
\caption{\small \label{fig:design_15}
Best function evaluation for RNA Design of puzzle $\#15$  with $n = 30$.}
\end{figure}
\begin{figure}[ht]
\center
\includegraphics[width=.9\linewidth]{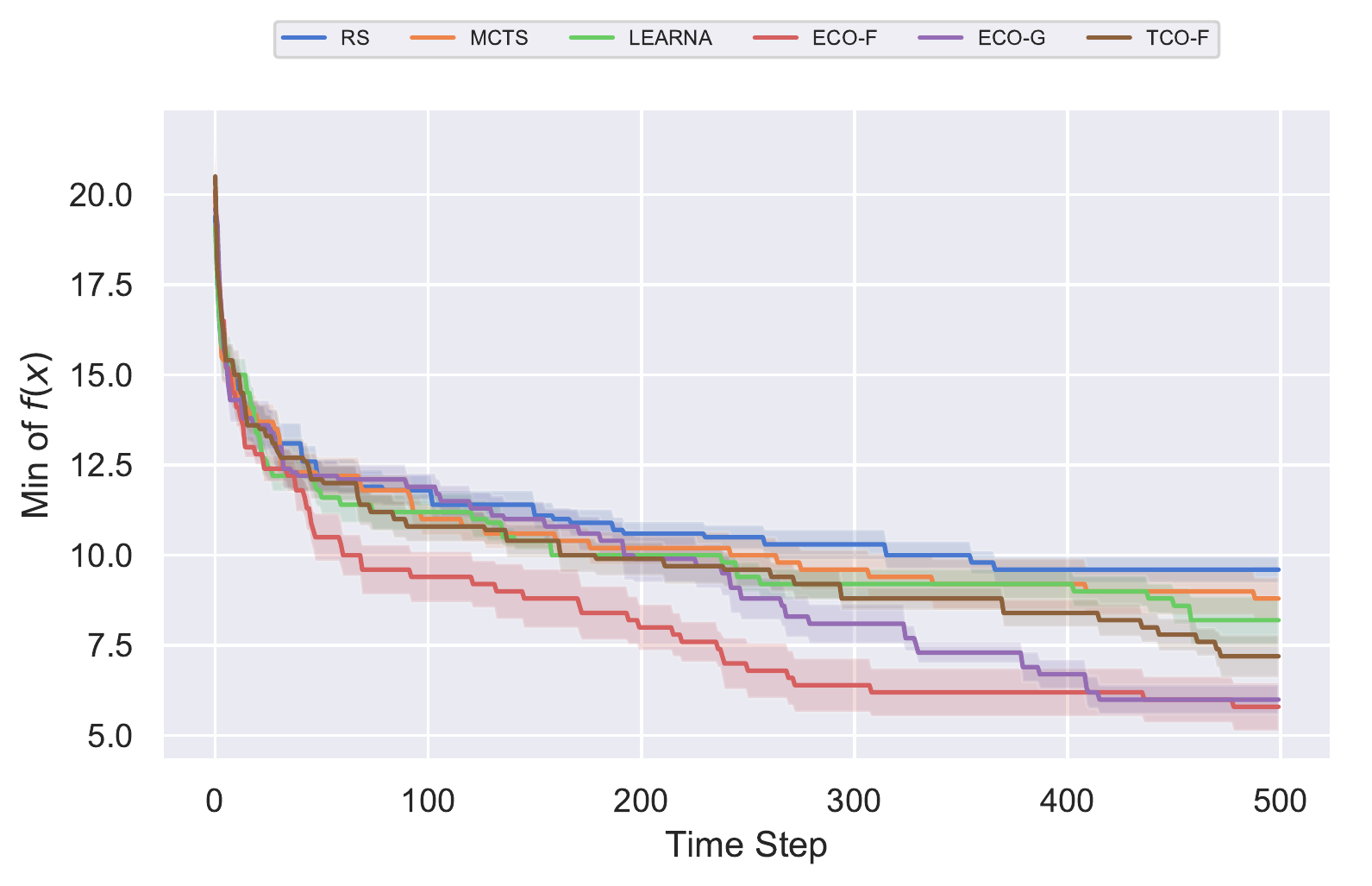}
\caption{\small \label{fig:design_41_app}
Best function evaluation for RNA Design of puzzle $\#41$  with $n = 35$.}
\end{figure}

For our experiments, we focus on three puzzles from the Eterna-100 dataset \cite{MODENA}. Two of the selected sequences (puzzles $15$ and $41$ of lengths $30$ and $35$, resp.), despite their fairly small lengths, are challenging for many algorithms (see \cite{MODENA}). In both puzzles, our MCTS variants (ECO-F and ECO-G) are able to significantly improve the performance of MCTS when limited number of true rewards (i.e. black-box evaluations) are available. All algorithms outperformed RS as expected. Within the given $500$ evaluation budget, TCO-F, ECO-G, and especially ECO-F, are superior to LEARNA by a substantial margin (see Figure \ref{fig:design_15}). In puzzle number $41$ (Figure \ref{fig:design_41_app}), again TCO-F, ECO-G and ECO-F significantly outperform LEARNA, over the given number of evaluations. Interestingly, ECO-F is able to outperform LEARNA throughout the evaluation process, and in average finds a far better final solution than LEARNA.

\begin{figure}[ht]
\center
\includegraphics[width=.9\linewidth]{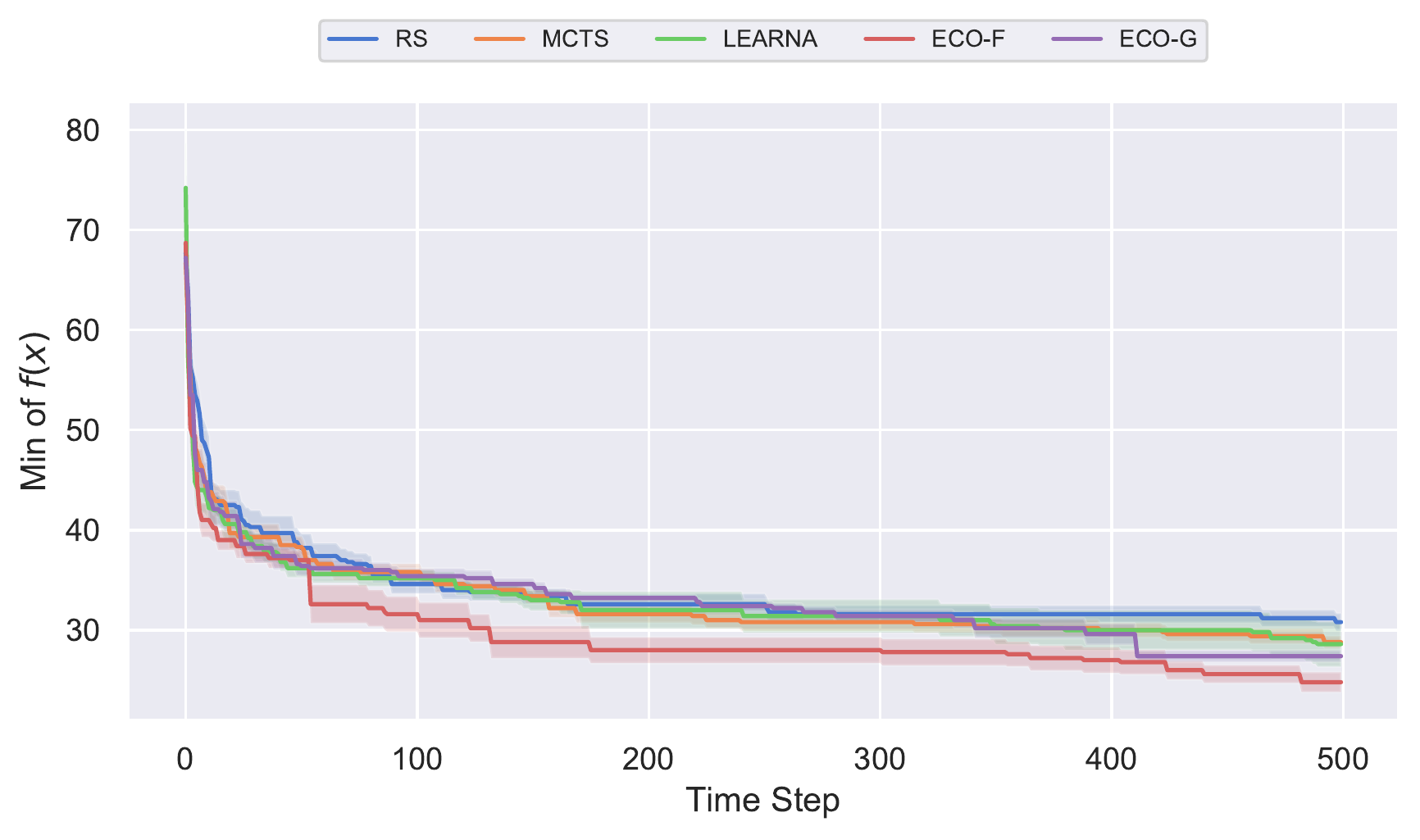}
\caption{\small \label{fig:design_70}
Best function evaluation for RNA Design of puzzle $\#70$  with $n = 184$.}
\end{figure}

The final sequence is puzzle $\#70$ of length $184$. The results of different algorithms over the latter puzzle is shown in Figure \ref{fig:design_70}. As we can see from this figure, MCTS-RNA and LEARNA perform very similarly over the given $500$ evaluation budget. ECO-F is able to outperform the remaining algorithm throughout the evaluation steps. Initially, ECO-G has a similar performance to those of MCTS-RNA and LEARNA, but offers an improved performance over the latter two just after $400$ steps. Due to the higher dimensionality of this puzzle, we drop TCO-F due to its high computation cost (expected time: $> 10$ days).

\section{Choice of AFO}
\label{app:choice_afo}

Throughout the experiments, we designated SA and MCTS as AFO for generic BBO and design problems, respectively. The latter choice was made in accordance with the literature, where SA is typically used as a baseline and method of choice for the generic BBO problem, whereas MCTS has been commonly used for the design problem. For instance, SA has been considered as a baseline and/or AFO in \cite{COMEX}, \cite{BOCS}, and \cite{COMBO} (albeit with a different algorithm than ours). On the other hand, MCTS (i.e. RNA-MCTS as well as its variations) is perhaps the most popular RNA design technique in the literature. Here, we point out that both SA and MCTS can be used for both generic BBO and design problems. In this section, we compare the performance of different AFO methods in each problem. 

\begin{figure}[ht]
\center
\includegraphics[width=.9\linewidth]{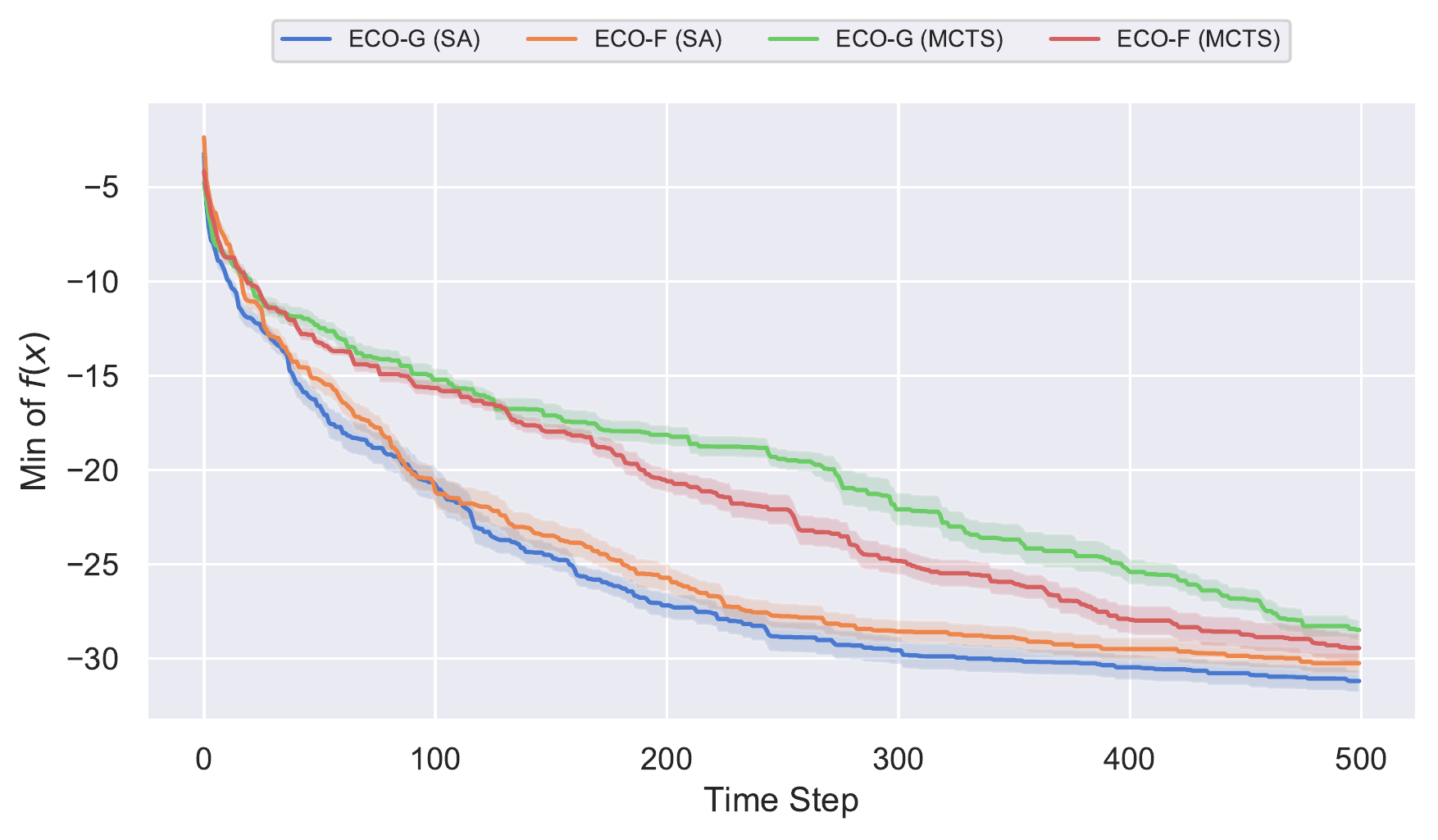}
\caption{\small \label{fig:acquisition_rna_optimization}
Comparison of different AFO methods for the generic BBO problem of RNA sequence optimization with $n = 30$.}
\end{figure}

First, we consider the generic BBO problem of RNA sequence optimization with $n=30$ (considered in Section \ref{sec:experiments}).
Figure \ref{fig:acquisition_rna_optimization} demonstrates the performance of ECO-F and ECO-G when SA or MCTS are used as AFO. As we can see from this figure, the SA-as-AFO variants perform slightly better than MCTS-as-AFO counterparts over $500$ steps. In particular, although the performance gap is initially moderately large, over time this performance gap becomes smaller.

\begin{figure}[ht]
\center
\includegraphics[width=.9\linewidth]{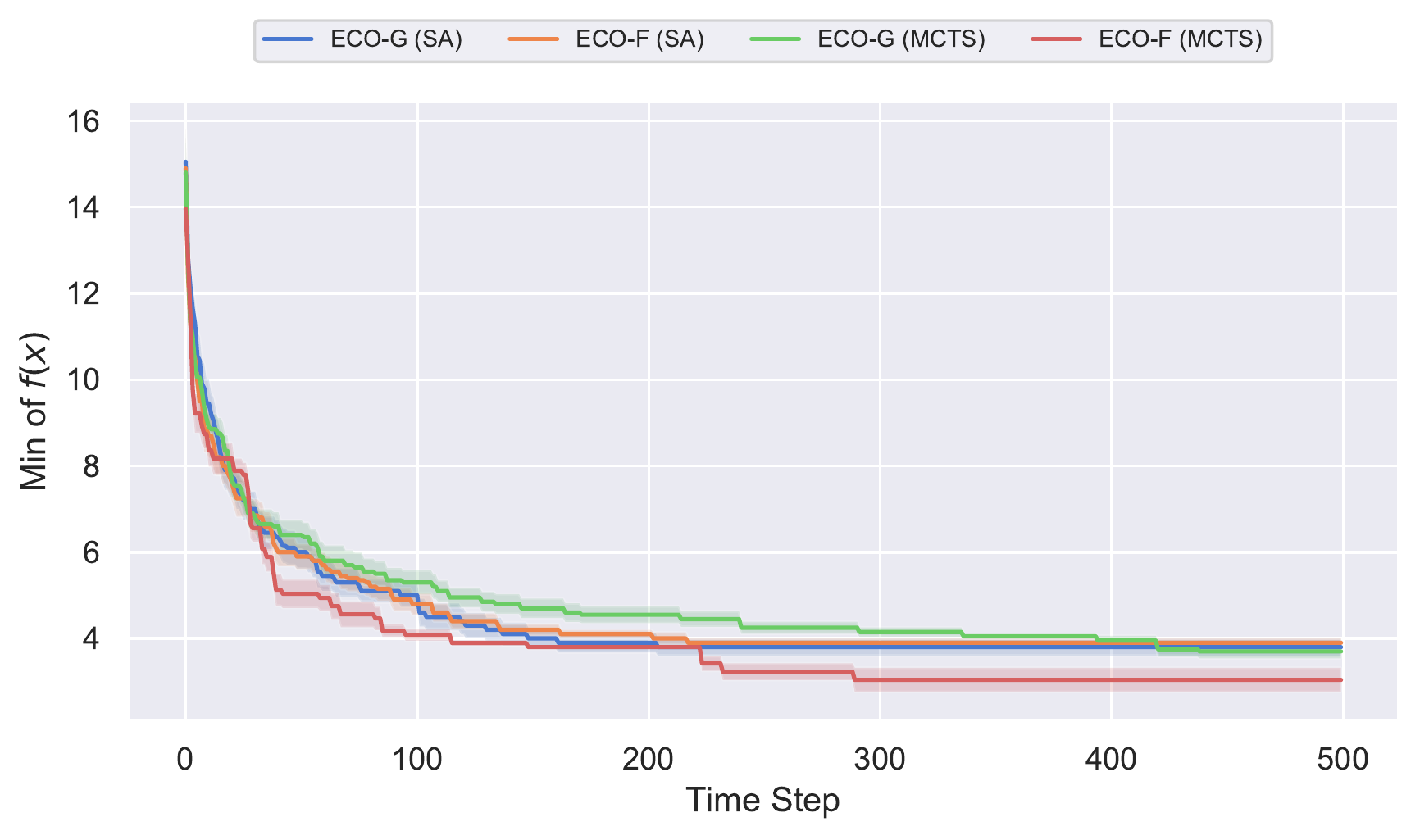}
\caption{\small \label{fig:acquisition_design_15}
Comparison of different AFO methods for design puzzle $\#15$ with $n = 30$.}
\end{figure}

\begin{figure}[ht]
\center
\includegraphics[width=.9\linewidth]{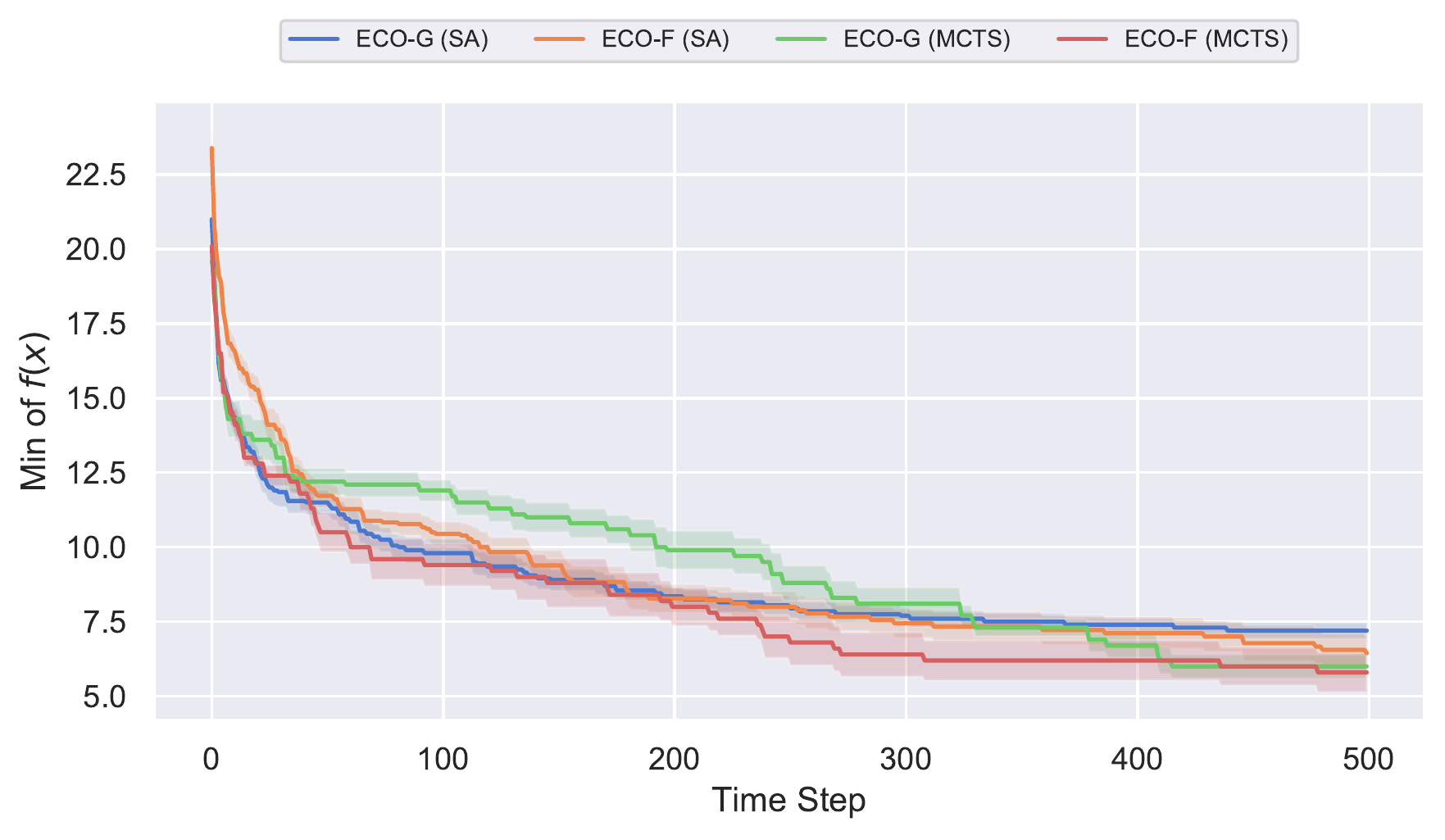}
\caption{\small \label{fig:acquisition_design_41}
Comparison of different AFO methods for design puzzle $\#41$ with $n = 35$.}
\end{figure}

Next, we consider two design problems considered in Section \ref{sec:experiments}: puzzles $\#15$ and $\#41$. Note that, when using SA as AFO, we apply the softmax operator (in Algorithm \ref{algo:SA}) over the set of $\{GC, CG, AU, UA\}$ if the corresponding variable is part of a paired base. As we can see from Figure \ref{fig:acquisition_design_15}, for puzzle $\#15$, ECO-F with MCTS outperforms the remaining algorithms. The rest of the algorithms have very similar performances, with ECO-F (MCTS) marginally surpassing the SA variants. As we can see from Figure \ref{fig:acquisition_design_41}, for puzzle $\#41$, the MCTS variant of ECO-F slightly outperforms its SA variant, whereas the MCTS variant of ECO-G maintains a bigger gap with its SA variant.

\section{Order of the Surrogate Model}
\label{app:order}

\begin{figure}[ht]
\center
\includegraphics[width=.9\linewidth]{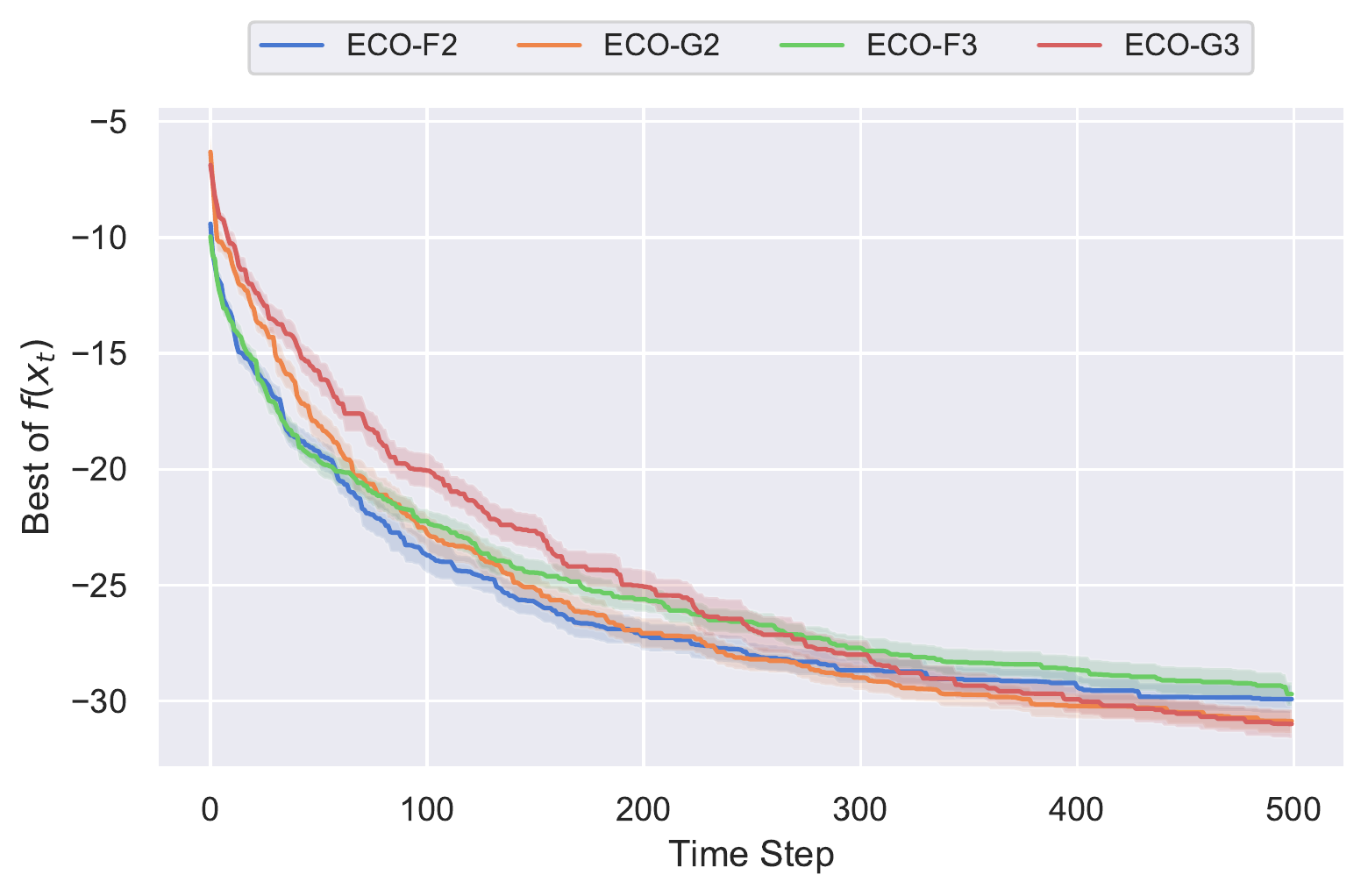}
\caption{\small \label{fig:rna_order}
Impact of model order on the performance of ECO-F/G in the RNA optimization problem with $n = 30$.}
\end{figure}

As mentioned in Section \ref{sec:experiments}, we used $m = 2$ as the maximum order of the representations in all the experiments. In this section, we focus on the generic BBO problem of RNA optimization and investigate the impact of the model order on the performance of the proposed ECO algorithms. In particular, we compare the performance of the algorithm at $m = 3$ with that of $m = 2$. As we can see from Figure \ref{fig:rna_order}, at smaller evaluation budgets, the order $2$ models moderately outperform the order $3$ counterparts in both ECO-F and ECO-G. As we increase the number of samples, this performance gap becomes smaller. At the $500$ evaluation budget, ECO-G3 outperforms ECO-G2 by a small margin of $0.1$. At the same evaluation budget, ECO-F3 is slightly inferior to ECO-F2 by a margin of $0.2$. Considering the convergence behavior of the curves at order $3$ versus those of order $2$, we expect the former models to eventually outperform the latter models at higher number of evaluations. 
However, since in BBO problems sample efficiency is typically of main concern, it would make sense to use low-order approximations. We point out that a similar observation was made in \cite{BOCS} for the Boolean case, where higher order models suffer from a slower start due to the higher dimensionality of the parameter space. From our experiments in categorical problems, this behavior seems to be even more pronounced due to the higher dimensionality of the categorical domains. Finally, we point out that in general a higher order increases the
expressiveness of the model but also decreases the accuracy of the predictions when data is limited (e.g. see chapter $14.6$ in \cite{gelman2013}).

A summary of the computation times for ECO algorithms at different model orders is given in table \ref{times_order}. Since the complexity of ECO is linear in the number of experts, which exponentially grows with the model order $m$, we observe an increase in the computational complexity of ECO-F3 (ECO-G3) versus that of ECO-F2 (ECO-G2) by a factor of $9.7$ ($16.3$).

\begin{table}[ht]
\caption{Average computation time per step (in Seconds) at different model orders.}
\label{times_order}
\begin{center}
\begin{small}
\begin{sc}
\begin{tabular}{lccr}
\toprule
ECO-F2 & ECO-F3 & ECO-G2 & ECO-G3 \\
\midrule
2.0  & 19.4  & 5.7  & 93.1  \\
\bottomrule
\end{tabular}
\end{sc}
\end{small}
\end{center}
\end{table}

\section{Impact of Categorical-Level Permutations}
\label{app:permutations}

It is easy to see that the abridged one-hot encoded Fourier representation is symmetric with respect to the categorical levels of different variables. In this section, we experimentally evaluate the impact of permutations of categorical levels in the group-theoretic Fourier representation since the impact of such permutations is not immediately obvious. In particular, we consider the following two problems: the Latin Square problem, and the RNA optimization problem. For the former problem, we already know that the problem is symmetric with respect to different categorical levels (e.g. swapping the roles of $2$ and $3$ has no impact on the solutions). The latter problem, on the other hand, is asymmetric with respect to different categorical levels (e.g. swapping the roles of $A$ and $G$ would likely change the solution(s) dramatically). 

For each problem, we consider four different permutations: the identity permutation $\sigma_0$ (i.e. the original assignment of categorical levels to Fourier characters), a fixed arbitrary permutation $\sigma_1$, an $n$-cyclic permutation $\sigma_2$, and a random permutation $\sigma_3$, where the permutation is selected uniformly at random per each run. In particular, $\sigma_1$ and $\sigma_2$ (in Cauchy's two-line notation) for the two problems, the Latin Square problem and the RNA optimization problem respectively, are given below:

\[
  \sigma_1 = \begin{pmatrix}
    0 & 1 & 2 & 3 & 4 \\
    3 & 2 & 4 & 0 & 1
  \end{pmatrix} \quad \quad
  \sigma_2 = \begin{pmatrix}
    0 & 1 & 2 & 3 & 4 \\
    1 & 2 & 3 & 4 & 0
  \end{pmatrix} 
\]
\[
  \sigma_1 = \begin{pmatrix}
    $A$ & $C$ & $G$ & $U$  \\
    $U$ & $A$ & $G$ & $C$
  \end{pmatrix} \quad \quad
  \sigma_2 = \begin{pmatrix}
    $A$ & $C$ & $G$ & $U$  \\
    $C$ & $G$ & $U$ & $A$
  \end{pmatrix} 
\]

For each permutation (and each problem), we compute the $95\%$ confidence interval around the mean over $20$ runs for the best $f(x_t)$ found until time step $500$. A summary of results for the two problems over different permutations is given in Table \ref{table:permutations}. As we can see from this table, for any given problem, the results of different permutations fall within the confidence intervals of one another. In addition, we see a very similar behavior in both problems, although the Latin Square problem is symmetric whereas the RNA optimization problem is asymmetric. Overall, from these results, we do not see any evidence that the performance of ECO-G (i.e. with the group-theoretic Fourier representation) depends on the permutations of categorical levels (i.e. the assignment of such levels to characters of the representation).


\begin{table}[t]
\caption{Impact of categorical level permutations on ECO-G.}
\label{table:permutations}
\begin{center}
\begin{small}
\begin{sc}
\begin{tabular}{lcccc}
\toprule
Permutation  &  Latin Square & RNA Optimization  \\
\midrule
 $\sigma_0$ & $0.82 \pm 0.39$  & $-30.40 \pm 1.19$ \\
 $\sigma_1$ & $0.60 \pm 0.36$ & $-30.68 \pm 1.49$ \\
 $\sigma_2$ & $0.70 \pm 0.40$ & $-31.44 \pm  1.64$ \\
 $\sigma_3$ & $0.94 \pm 0.42$ & $-30.88 \pm 1.35$ \\
\bottomrule
\end{tabular}
\end{sc}
\end{small}
\end{center}
\end{table}

\section{Comparison with Batch Optimization}
\label{app:batch}

As pointed out in the related work section, batch and sequential settings are fundamentally different. Despite the differences, in order to gauge the overall quality of our algorithm, we ran extensive experiments to test and compare our algorithms over two RNA landscapes of \cite{sinai2021adalead} with state-of-the-art batch optimization methods. We run our algorithms sequentially from steps $1$ to $500$, and batch algorithms over $5$ batches, each of size $100$. For $\texttt{L14-RNA1}$ and $\texttt{L14-RNA1+2}$ landscapes with sequence length of $14$, the results are given at $100$, $200$, $300$, $400$, and $500$ samples averaged over $10$ runs. Note that this is a maximization problem, where the max value is approximately equal to $1$ according to \cite{sinai2021adalead}. Comparison is done against AdaLead \cite{sinai2021adalead}, DyNAPPO \cite{Angermueller2020Model}, CbAS \cite{brookes2019conditioning}.

\begin{table}[!htbp]
\caption{Results for $\texttt{L14-RNA1}$ Landscape.}
\label{first_landscape}
\begin{center}
\begin{small}
\begin{sc}
\begin{tabular}{lccccccr}
\toprule
$\#$ of Samples & $100$ & $200$ & $300$ & $400$ & $500$ \\
\midrule
ECO-F & $0.82$ & $0.861$ & $0.879$ & $0.91$ & $0.92$ \\
ECO-G & $0.842$ & $0.9$ & $0.934$ & $0.952$ & $0.96$ \\
TCO-F & $0.925$ & $0.987$ & $0.995$ & $0.998$ & $0.999$ \\
AdaLead & $0.601$ & $0.707$ & $0.799$ & $0.875$ &  $0.912$ \\
DyNA-PPO & $0.599$ & $0.708$ & $0.739$ & $0.771$ & $0.782$ \\
CbAS & $0.615$ & $0.657$ & $0.711$ & $0.715$ & $0.715$ \\
\bottomrule
\end{tabular}
\end{sc}
\end{small}
\end{center}
\end{table}

\begin{table}[!htbp]
\caption{Results for $\texttt{L14-RNA1+2}$ Landscape.}
\label{second_landscape}
\begin{center}
\begin{small}
\begin{sc}
\begin{tabular}{lccccccl}
\toprule
$\#$ of Samples & $100$ & $200$ & $300$ & $400$ & $500$ \\
\midrule
ECO-F & $0.823$ & $0.876$ & $0.897$ & $0.906$ & $0.922$ \\
ECO-G & $0.818$ & $0.897$ & $0.953$ & $0.963$ & $0.963$ \\
TCO-F & $0.947$ & $0.99$ & $0.995$ & $0.998$ & $0.998$ \\
AdaLead & $0.61$ & $0.78$ & $0.88$ & $0.92$ & $0.92$ \\
DyNA-PPO & $0.67$ & $0.72$ & $0.76$ & $0.78$ & $0.79$ \\
CbAS & $0.52$ & $0.60$ & $0.62$ & $0.64$ & $0.65$ \\
\bottomrule
\end{tabular}
\end{sc}
\end{small}
\end{center}
\end{table}

As we can see from the results in Tables \ref{first_landscape} and \ref{second_landscape}, our algorithms, particularly TCO-F, outperform the batch optimization counterparts with a large margin. In particular, $100$ samples with TCO-F are sufficient to outperform AdaLead with $500$ samples ($0.925$ versus $0.912$).

\section{Comparison between TCO-F and BOCS+Vanilla One-hot encoding}
\label{sec:BOCS_comparison}

BOCS \cite{BOCS} is devised to address problems over the Boolean hypercube as it uses a Boolean Fourier representation as its surrogate model. Experimentally, they also focused on the Boolean case exclusively. They have mentioned in the appendix that it can be extended to problems over categorical variables via vanilla one-hot encoding (but again with no experiments). However, vanilla one-hot encoding suffers from the curse of dimensionality due to the large number of one-hot variables. Our abridged one-hot encoded Fourier representation addresses this problem by reducing the number of variables significantly. The TCO-F algorithm, while it is based on the principles underlying BOCS, uses the abridged one-hot encoded representation that addresses the categorical problems in a computationally efficient way.

\begin{figure}[ht]
\center
\includegraphics[width=\linewidth]{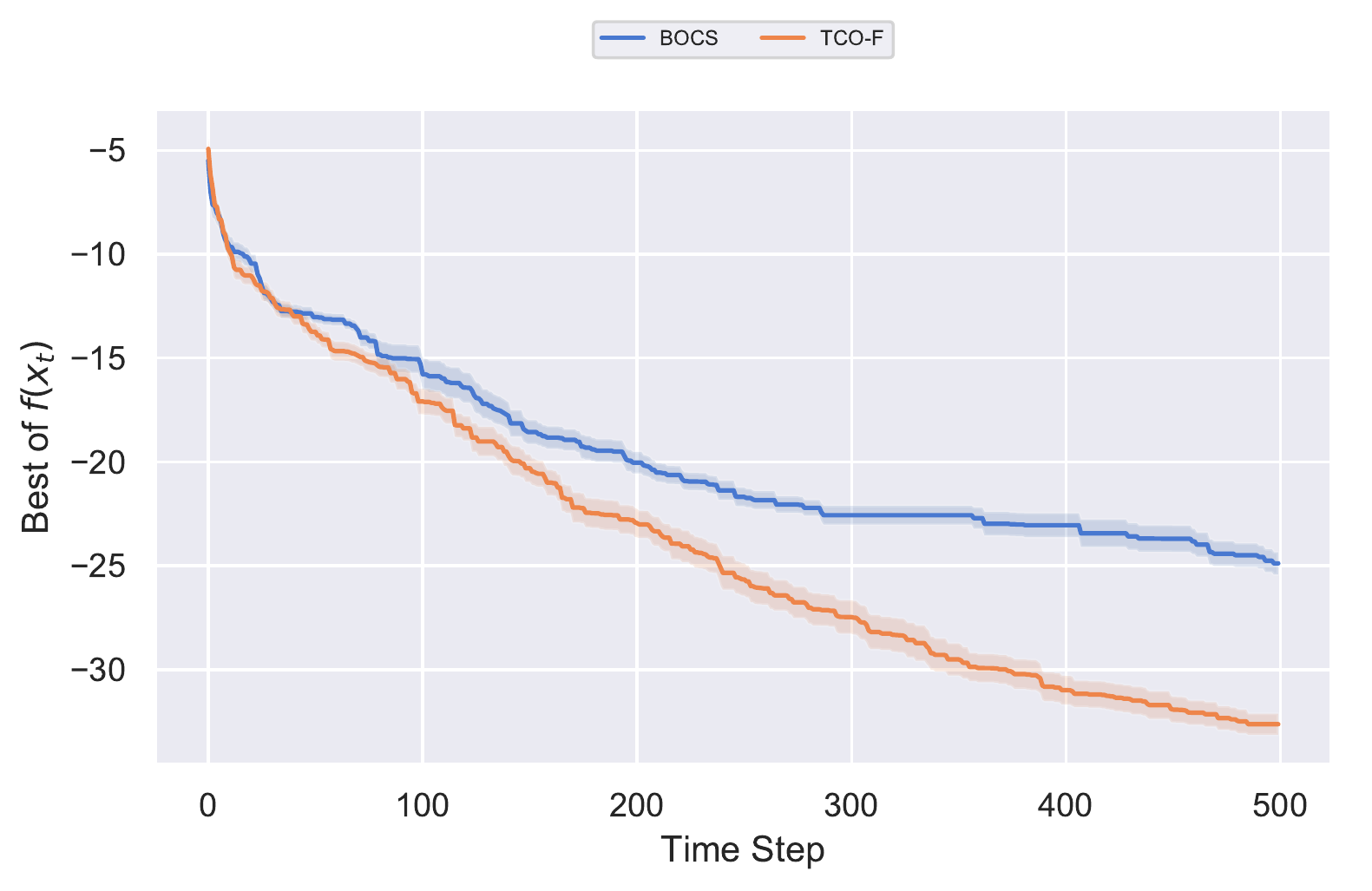}
\caption{\small \label{fig:bocs_vanilla}
Comparison of TCO-F with BOCS+Vanilla one-hot encoding over the RNA sequence optimization problem.}
\end{figure}

In this section, we compare the performance of our TCO-F against BOCS with vanilla one-hot encoding, as suggested in \cite{BOCS}. Figure \ref{fig:bocs_vanilla} compares the performance of TCO-F and BOCS (+ vanilla one-hot encoding) over $500$ time steps. As we can see from this figure, BOCS's best result until $150$ steps is $-18.3$ whereas TCO-F achieves $-20.0$. After $500$ steps, BOCS+vanilla one-hot encoding achieves $-25.0$ whereas our TCO-F algorithm achieves a much better solution of $-32.6$.
From a computational standpoint, each step of BOCS+vanilla one-hot encoding on average takes $651$ seconds to process whereas our TCO-F algorithm only takes $67$ seconds on average for each time step. 

\section{Existing Assets}

In out experiments, we compared our results against COMBO \cite{COMBO} and LEARNA \cite{runge2018learning}:

\subsection{COMBO}
 
Codes: \url{https://github.com/tsudalab/combo}

License: This package is distributed under the MIT License.

\subsection{LEARNA}

Codes: \url{https://github.com/automl/learna}

License: Licensed under the Apache License, Version $2.0$
\\

In addition, we used the following existing datasets:

\subsection{RNA Problem}

Package: \url{https://www.tbi.univie.ac.at/RNA/}

License: The programs, library and source code of the Vienna RNA Package are free software.  

\subsection{Pest Control Problem}

We used the problem as included in the COMBO repository:

Codes: \url{https://github.com/tsudalab/combo}

License: This package is distributed under the MIT License.

\end{document}